\documentclass{article}

\usepackage[preprint]{corl_2026}
\usepackage{amsmath}
\usepackage{amssymb}
\usepackage{booktabs}
\usepackage{graphicx}
\usepackage{xcolor}
\usepackage{xspace}
\usepackage{wrapfig}
\usepackage{enumitem}
\usepackage{gradient-text}
\usepackage{caption}

\title{Tactile Genesis: Exploring Tactile Sensors \\ at Scale for Learning Dexterous Tasks}

\author{
  Trinity Chung$^{\dagger,1}$ \quad Kashu Yamazaki$^{1}$ \quad Dhruv Patel$^{1}$ \\
  \bfseries Alexis Duburcq$^{2}$ \quad Yiling Qiao$^{2}$ \quad Katerina Fragkiadaki$^{1}$ \quad Aran Nayebi$^{1}$ \\[2pt]
  \normalfont
  $^{1}$Carnegie Mellon University \quad $^{2}$Genesis AI 
  \\
  $^{\dagger}$Corresponding Author:
  \texttt{trinityc@cmu.edu} \\
  \normalfont
}

\newcommand{\ours}{\gradientRGB{Tactile Genesis}{50,0,0}{200,0,0}\xspace}
\newcommand{\celsius}{\ensuremath{^{\circ}C}\xspace}

\begin{document}
\maketitle

\begin{abstract}
Tactile sensing is critical for contact-rich dexterous manipulation, yet it remains unclear which tactile abstractions a policy needs and when richer tactile fields justify their hardware cost.
This is hard to study empirically: each sensor effectively defines a new robot, and no lab can replicate the same learning experiment across all of them.
We present \ours, a GPU-parallel tactile sensor simulation platform that exposes binary contact, contact depth, per-taxel kinematic force/torque, elastomer marker displacement, geometry-aware proximity, contact audio, and a voxelized temperature field (the first of its kind in robot learning physics simulation platforms) under a common interface, with configurable placement, resolution, and a realistic noise model (drift, hysteresis, dead taxels, crosstalk).
It scales past 20,000 parallel environments and 1,000 taxels on a single GPU, improving throughput by 3 to 20 times over previous tactile simulators.
We train teacher-student policies on three dexterous tasks, ablating sensor type, placement, resolution, and noise, and verify transfer to the real XHand1.
Proprioception alone is insufficient on every task.
Sensor placement dominates sensor type: fingertip-only coverage trails whole-hand coverage by a wide margin, while adding the palm and proximal phalanges closes most of the gap to the privileged teacher.
Resolution matters far less than coverage: placing 200 taxels across the whole hand suffices across tasks.
We find that force/torque per taxel is consistently the most useful sensor type.
These results give concrete guidance for both future tactile hardware design for improving robot hands and policy-side observation choice in dexterous manipulation.
\url{https://neuroagents-lab.github.io/tactile-genesis/}
\end{abstract}


\begin{figure}[h]
    \centering
    \includegraphics[width=\linewidth]{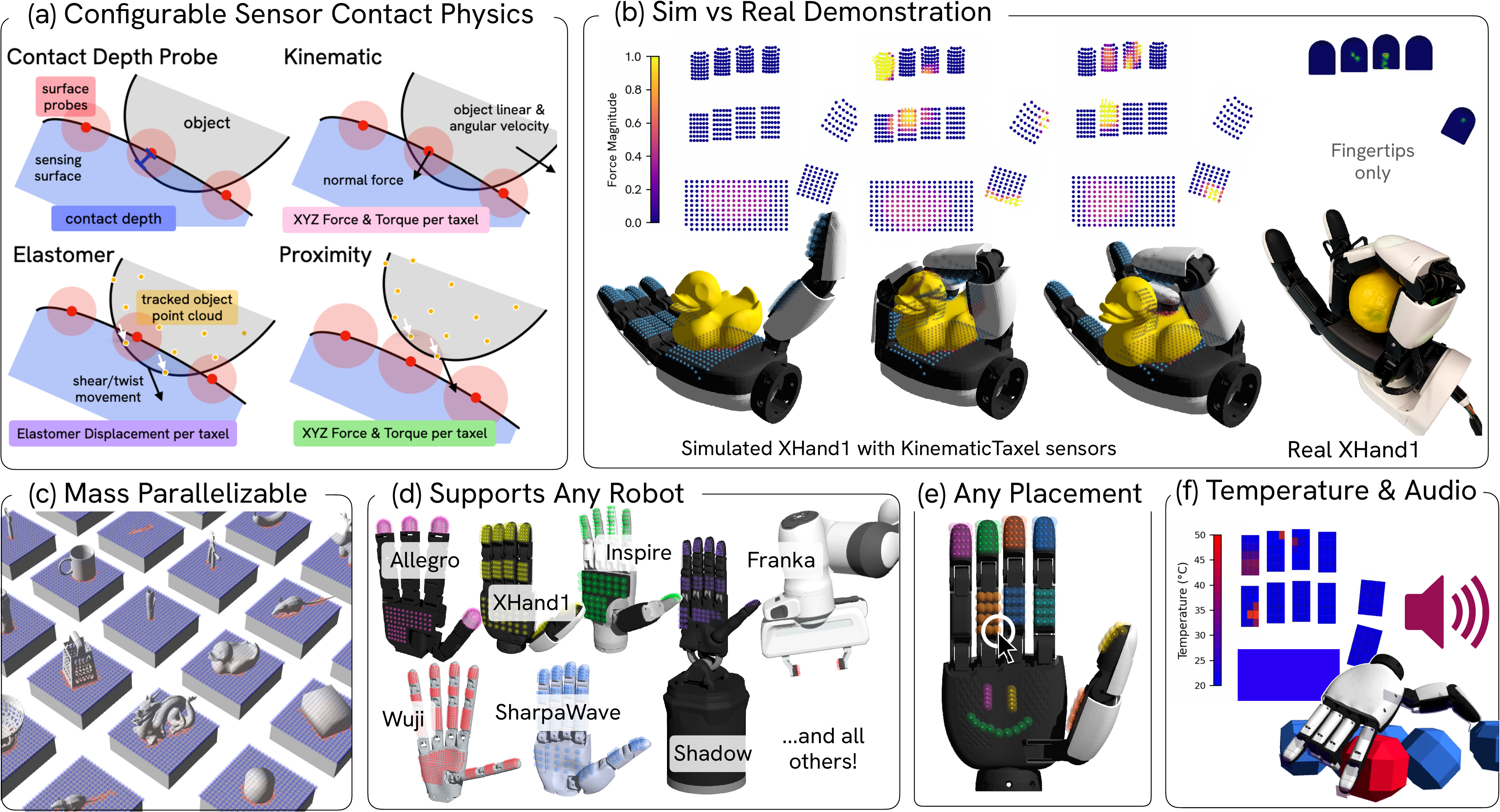} 
    \caption{
    \textbf{Overview of \ours features.}
    (a) The sensor physics can be configured to match their real sensor analogues, including 6-axis force/torque measurements, elastomer displacement, and proximity signal.
    (b) A visual comparison of the simulated tactile force reading per taxel on an XHand1 compared to the real XHand1's sensor fidelity. 
    (c) Our sensor implementations are highly parallelized and supports heterogeneous objects and randomization.
    (d) The simulated tactile sensors can be applied on any robot hardware surface and (e) the placement can be of arbitrary shape and resolution.
    (f) A temperature sensor which simulates contact heat transfer, heat diffusion, heat generation, and radiation, and a contact audio sensor which outputs high frequency signals based on material properties which the rigid body physics engine alone cannot capture.
    }
    \label{fig:overview}
\end{figure}

\section{Introduction}

Dexterous manipulation is fundamentally contact-rich. Vision can localize objects before contact, and proprioception can track the robot's own motion, but many manipulation failures happen through local phenomena that neither modality observes directly: slip, incipient loss of force closure, decoupled object-hand motion, small contact timing errors, and hidden contacts inside clutter. Biological manipulation makes the same point from the other direction: humans and animals can manipulate objects with little or no visual feedback once contact is established. The question for robot learning is therefore not whether touch can help, but what kind of tactile information a policy needs.

Current tactile hardware spans capacitive arrays~\citep{song2026fingertip}, magnetic skins~\citep{bhirangi_reskin_2021}, vision-based elastomer sensors~\citep{yuan_gelsight_2017}, strain gauge~\citep{xu2025multimodaltactilefingertipdesign}, contact microphones~\citep{xu2025multimodaltactilefingertipdesign,mejia2024hearing}, multisensory fingertips~\citep{,chelly_tactile-based_2025,kitouni_fingertip_2024,higuera_tactile_2025}. These sensors differ in spatial resolution, bandwidth, cost, durability, wiring complexity, calibration burden, and suitability to cover the entire hand versus only the fingertip. A lab that buys or builds one sensorized hand usually cannot repeat the same dexterous learning experiment across all of these alternatives.

Simulation gives us the missing controlled comparison, but only if the simulated tactile sensing is realistic enough to transfer and fast enough for large-scale policy training. Previous work has already shown that carefully calibrated tactile simulators can support sim-to-real manipulation for visuotactile sensors~\citep{zhao_fots_2024,su_tacmap_2026,dang_hydroshear_2026}. \textbf{We ask a complementary design question:} if we can simulate many tactile abstractions at scale, which representation of touch is sufficient to learn general-purpose dexterous manipulation tasks?

We present \ours, a scalable and hardware-agnostic platform for tactile sensor simulation. 
\ours covers a representative set of contact abstractions under a unified interface (Fig.~\ref{fig:overview}): binary contact, raw contact depth, per-taxel kinematic force/torque, elastomer marker displacement, and geometry-aware proximity.
Each sensor exposes configurable placement, resolution, and noise parameters (drift, hysteresis, dead taxels, crosstalk).
The implementation is GPU-parallelized across thousands of environments and taxels, making it usable as the tactile front-end for dexterous reinforcement learning (Fig.~\ref{fig:method_sensor_benchmark}).
Using this platform, we train teacher-student policies on three dexterous tasks and ablate sensor type, placement, resolution, and noise.
Our findings give concrete guidance for which tactile abstractions are worth the hardware cost on a given manipulation task. In summary, our contributions are:
\begin{enumerate}[leftmargin=*]
\item We introduce \ours, a GPU-parallel tactile simulation platform that unifies diverse tactile sensing abstractions under a common configurable interface, scaling to over 20,000 parallel environments and 1,000+ taxels on a single GPU.

\item  We implement a temperature sensor, the first in any robot learning simulation, and use it to learn a policy that locates a hot object among geometrically identical distractors from proprioception and temperature alone.
We ablate the thermal properties of the sensing surface and show that the low sensitivity of current real hardware is insufficient for learning this task.

\item We perform a controlled study of tactile representations across dexterous tasks, robot hands, sensor placements, resolutions, and noise settings.

\item We show that tactile placement matters more than sensor type, with whole-hand coverage substantially outperforming fingertip-only sensing, and that per-taxel force/torque is a strong default representation.
\end{enumerate}

\begin{figure}[t]
    \centering
    \includegraphics[width=\linewidth]{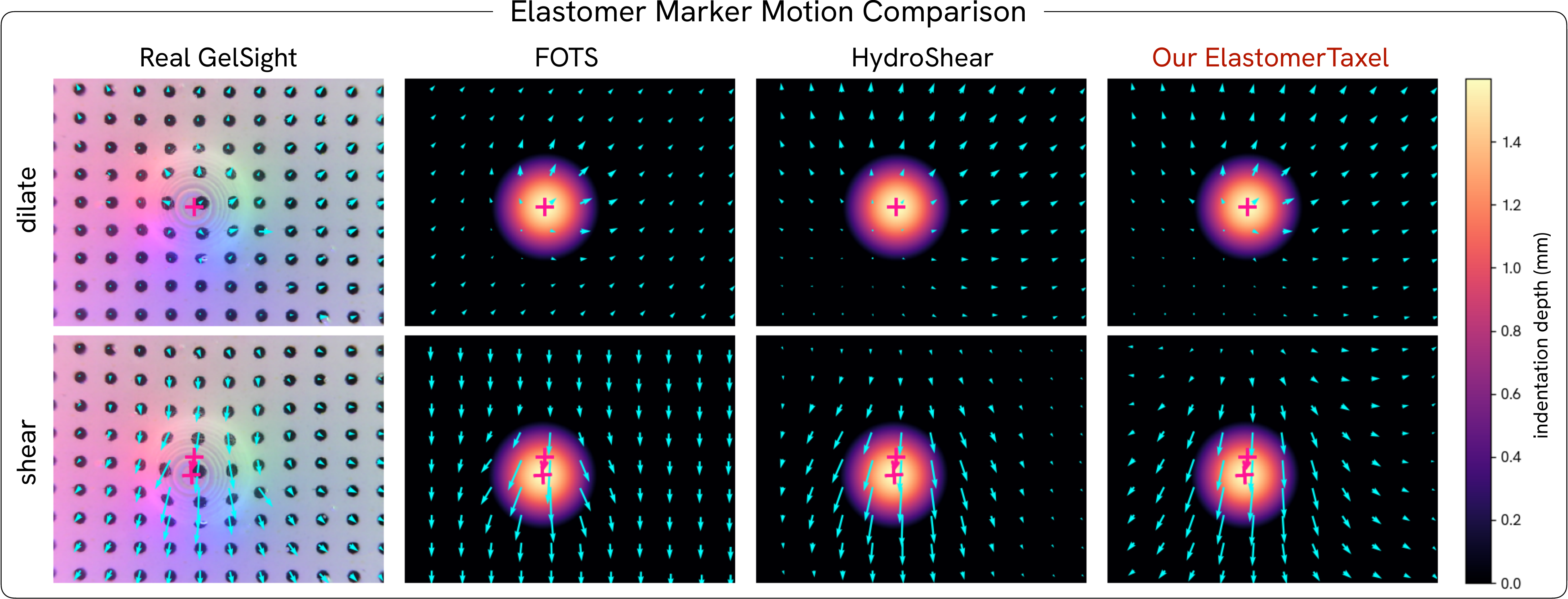}
    \vspace{4pt}

    \begin{minipage}[c]{0.66\linewidth}
        \caption{
        \textbf{Elastomer marker motion comparison.}
        Marker displacement fields on a real GelSight under dilation (normal indentation) and shear (tangential drag), compared with FOTS~\citep{zhao_fots_2024}, HydroShear~\citep{dang_hydroshear_2026}, and our \texttt{ElastomerTaxel}.
        The table on the right reports the relative marker-displacement error for each simulator after optimizing parameters to match the real image.
        The real GelSight image was obtained from the FOTS paper codebase, and we replicate the setup in sim, using our \texttt{ContactDepthProbe} sensor to measure depth.
        }
        \label{fig:elastomer_marker_comparison}
    \end{minipage}
    \hfill
    \begin{minipage}[c]{0.30\linewidth}
        \centering
        \footnotesize
        \setlength{\tabcolsep}{3pt}
        \begin{tabular}{@{}lccc@{}}
            \toprule
            \multicolumn{4}{c}{Relative RMSE ($\downarrow$)} \\
            \midrule
             & FOTS & HydroShear & Ours \\
            \midrule
            dilate & 0.514 & 0.403 & \textbf{0.329} \\
            shear  & 0.210 & 0.217 & \textbf{0.174} \\
            \bottomrule
        \end{tabular}
    \end{minipage}
\end{figure}

\section{Related Work}

\paragraph{Tactile hardware.}
Tactile hardware for robot hands spans a diverse set of transduction principles, each with its own resolution, bandwidth, footprint, and cost.
Capacitive arrays~\citep{song2026fingertip} and strain-gauge fingertips~\citep{xu2025multimodaltactilefingertipdesign} provide direct force readings but are typically restricted to the fingertip.
Magnetic skins such as ReSkin~\citep{bhirangi_reskin_2021} estimate deformation indirectly from magnetometer displacements and can cover larger areas at lower spatial resolution.
Vision-based elastomer sensors such as GelSight~\citep{yuan_gelsight_2017} resolve fine surface deformation but are bulky and again confined to flat pads.
Multimodal fingertips~\citep{higuera_tactile_2025,chelly_tactile-based_2025,kitouni_fingertip_2024} combine several of these channels at the cost of substantial wiring and calibration.
As a result, comparing sensors apples-to-apples on the same robot hand and task is rarely feasible in hardware: each sensor effectively defines a different hand.

\paragraph{Tactile simulation.}
Tactile simulators broadly fall into two families: full deformable physics, typically using the Finite Element Method (FEM), and rigid-body simulation with soft-contact post-processing.
FEM has historically been considered too slow for robot learning, but Taccel~\citep{li_taccel_2025} narrows this gap by combining Incremental Point Contact (IPC), deformable only on the sensor surface, with GPU parallelization, scaling to 4096 environments.
The rigid-body family extracts tactile signal from existing contact queries.
Tacmap~\citep{su_tacmap_2026} casts rays from the sensor surface to produce per-pixel contact depth, which is a good match for rigid-pad fingertips such as SharpaWave and XHand1.
We support the same query via either raycasting or a Signed Distance Function (SDF), and benchmark against Tacmap in Fig.~\ref{fig:prior_simulator_comparison}.
FOTS~\citep{zhao_fots_2024} simulates the visual output of GelSight-style sensors directly, bypassing the deformation physics by modeling how the elastomer indentation maps to the camera image.
HydroShear~\citep{dang_hydroshear_2026} extends this idea to GPU-parallel marker displacement, combining SDF-based depth with anchored point-cloud tracking to capture shear and twist.
Our analogous Elastomer sensor builds on HydroShear with substantially better throughput and memory usage (Fig.~\ref{fig:prior_simulator_comparison}).
TacSL~\citep{akinola2025tacsl} is likewise GPU-parallel, integrated into Isaac Lab, and renders a tactile depth image, RGB image, and a penalty-based tactile force field; we benchmark our throughput against it in Fig.~\ref{fig:prior_simulator_comparison}.
We do not render a tactile RGB image directly, since its appearance is specific to the sensor vendor (e.g. GelSight), but it can be derived from our \texttt{ContactDepthProbe} depth with an example-based renderer such as Taxim~\citep{si2022taxim}.
\citet{yin_learning_2025} uses sampled point-cloud tracking, not for deformation but to estimate ReSkin~\citep{bhirangi_reskin_2021} magnetometer readings; we include a corresponding Proximity sensor variant as well.
To our knowledge, no prior simulator offers all five of these abstractions (binary contact, depth, kinematic force/torque, elastomer displacement, proximity) under a common interface, nor does it have temperature sensors, which is what makes the representation ablation in this paper possible.

\paragraph{Tactile representations.}
What tactile information a policy actually needs is still an open empirical question.
\citet{miller_enhancing_2025} report that sparse binary contacts are sufficient for several in-hand manipulation skills.
Sparsh-X~\citep{higuera_tactile_2025} goes the other direction and fuses tactile images, audio, motion, and pressure to encode object-level physical properties.
ManiWAV~\citep{liu2025maniwav} shows that a contact audio sensor embedded in the gripper lets the robot learn contact modes and surface materials.
In between sit force/torque arrays and elastomer displacement fields, which expose richer local mechanics than binary contact without the bandwidth and calibration cost of full multimodal fingertips.
Prior work typically commits to a single point in this design space because the underlying hardware does.
We instead hold the policy architecture, task, and hand fixed and vary the tactile abstraction directly in simulation, asking which abstraction is sufficient on which task.

\begin{table}[ht]
    \centering
    \small
    \begin{tabular}{l p{0.6\linewidth} c}
        \toprule
        Sensor Type & Data Description & Shape \\
        \midrule
        Surface Distance Probe & Shortest distance to the surface of tracked object. & $(N)$ \\
        Contact Depth Probe & Raw contact depth scalar from simulated physics, either from SDF or raycasting and sphere-triangle intersection. & $(N)$ \\
        Contact Probe & Binarized contact with depth threshold and hysteresis & $(N)$ \\
        Kinematic Taxel & Per-taxel force/torque estimate from SDF depth, contact normal, and linear/angular velocity of object in contact. & $(N, 6)$ \\
        Proximity Taxel & Measures object surface mass within sensing distance using point cloud sampled on tracked objects for signal strength. Computes force/torque using object velocities. & $(N, 6)$ \\
        Elastomer Taxel & Marker displacement modeling the sensing surface as an elastomer. & $(N, 3)$ \\
        \midrule
        Temperature Grid & Temperature in \celsius over voxelized link. & $(N)$ \\
        Contact Audio & Block of $K$ synthesized vibration samples per step. & $(N, K)$ \\
        \bottomrule
    \end{tabular}
    \vspace{4pt}
    \caption{\textbf{\ours sensor implementations.} $N$ represents the number of probes/taxels.
    The full implementation details and configurable parameters per sensor are available in the Appendix.}
    \label{tab:genesis_sensor_classes}
\end{table}

\section{Method}

\subsection{Tactile Sensor Simulation}

\begin{figure}[ht]
    \centering
    \includegraphics[width=\linewidth]{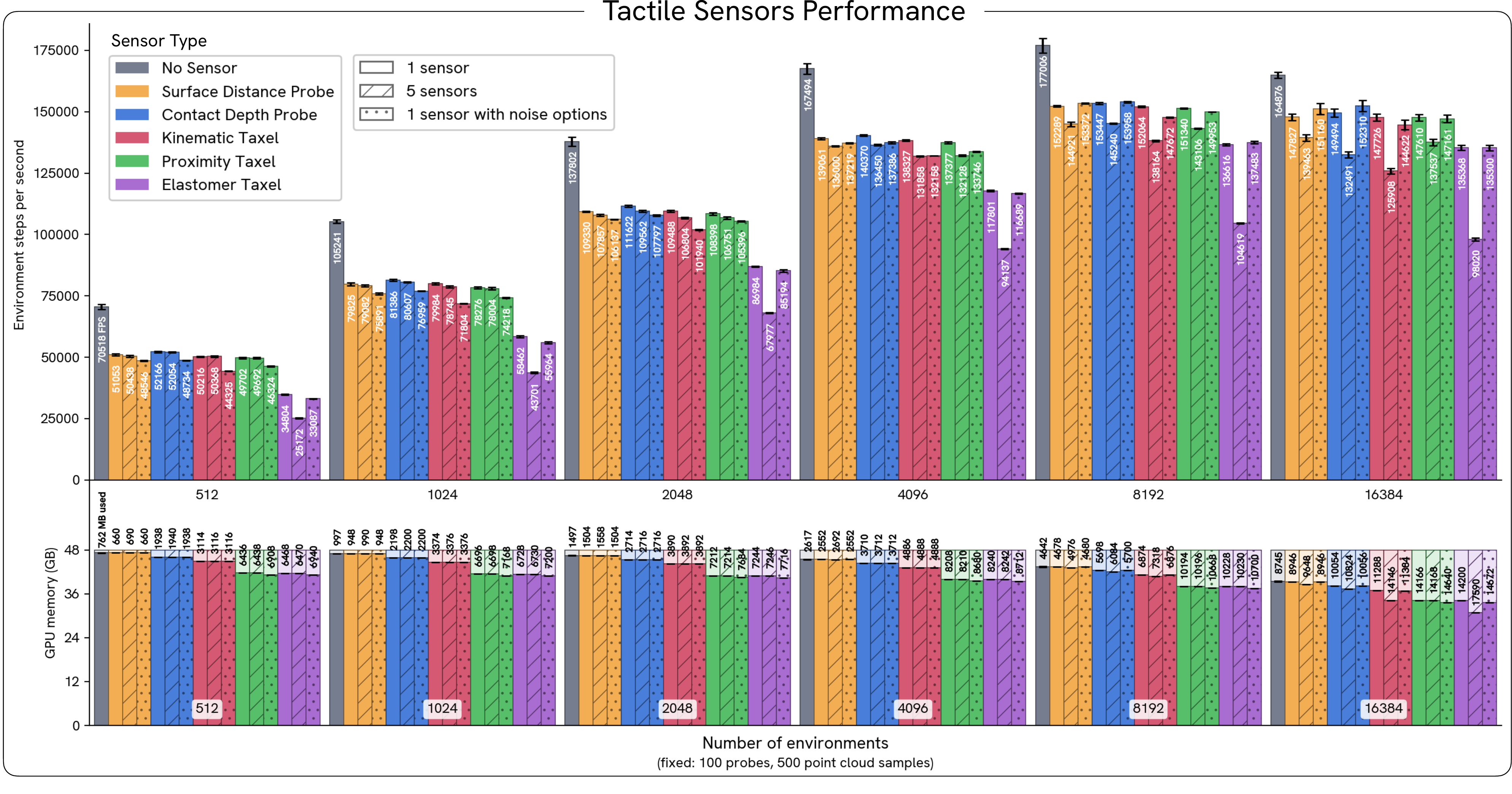}
    \caption{
    \textbf{Performance benchmark per each simulated sensor type.}
    We demonstrate that our sensors are able to be parallelized on a single NVIDIA RTX A6000 beyond 16,384 environments, with a total throughput of 150,000 environment steps per second (FPS). To isolate the effect of the sensors, we perform the benchmark in a simple scene of a pyramid of 10 cubes. We compare the performance of different sensor types, fixing the sensor resolution at $10\times 10$ (100 taxels) and 500 tracked point cloud samples (applicable to Elastomer and Proximity sensors). Adding more sensors or noise parameters adds small overhead (-10\% FPS) for most sensors. The Elastomer sensor slows as more sensors are added (-22\% FPS) since local displacement effects must be computed per sensor.
    }
    \label{fig:method_sensor_benchmark}
\end{figure}

\begin{figure}[ht]
    \centering
    \includegraphics[width=1.0\linewidth]{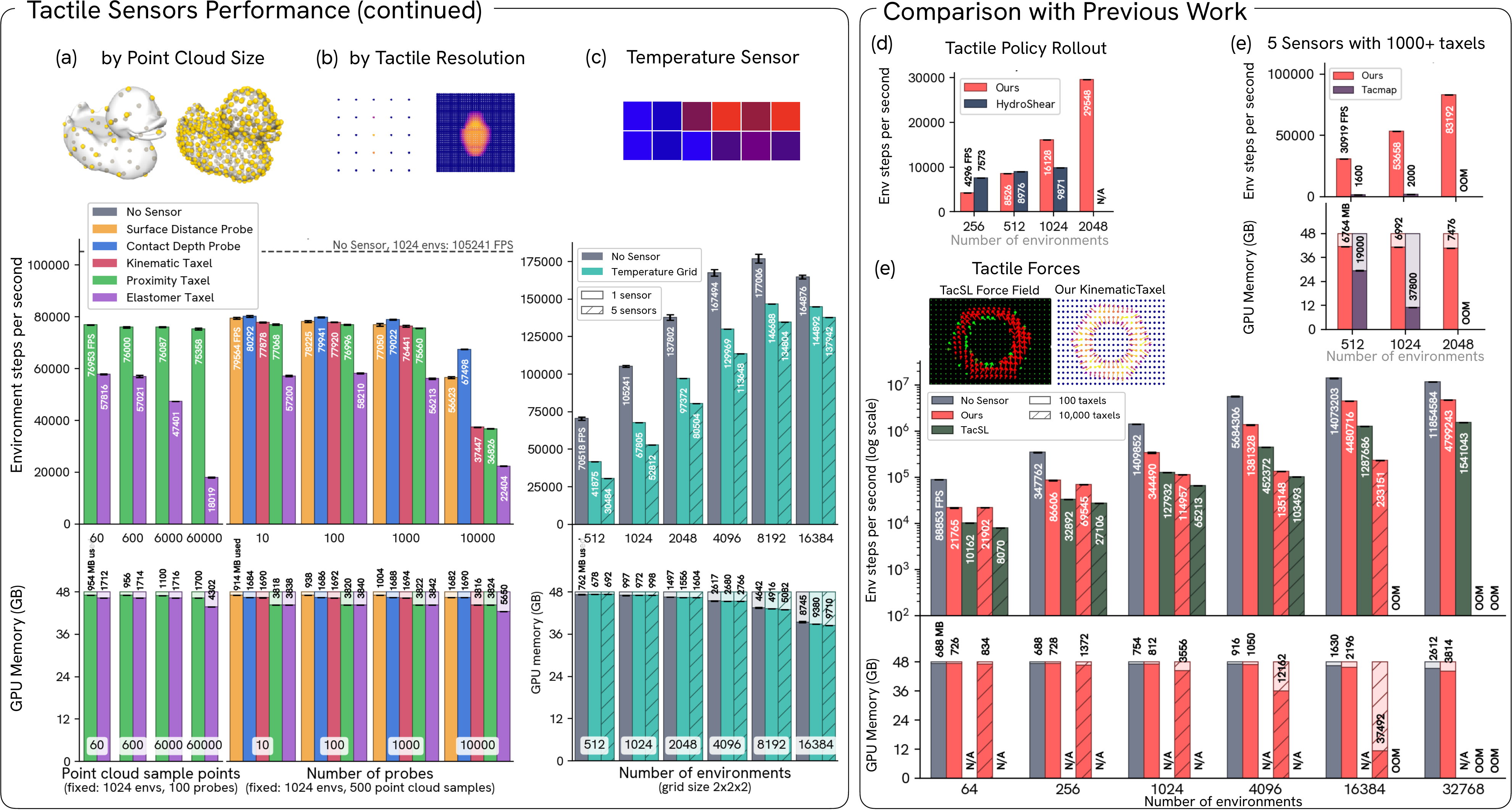}
    \caption{
    \small
    \textbf{Performance benchmark (continued).}
    (a, b) With the number of environments fixed at 1024 and varying point-cloud size and taxel count, our sensors retain low GPU memory and high throughput up to over 10,000 taxels per hand. The elastomer sensor has to track the motion of each object point in contact and therefore scales less well with point-cloud size, but point clouds larger than $\sim$6000 are not typically needed for dexterous tasks.
    (c) Our temperature sensor also scales well with number of environments, achieving 80\% of the no sensor baseline throughput FPS with 5 active sensors with 8 voxels each.
    \\
    \textbf{Comparison with previous work.} Comparison to performance numbers reported by Tacmap~\cite{su_tacmap_2026} and HydroShear~\citep{dang_hydroshear_2026}; neither paper reports the GPU used or metrics beyond 1024 environments. All \ours benchmarks were run on one NVIDIA RTX A6000.
    (d) Tacmap is based on SharpaWave, which has $>1,000$ tactile pixels per fingertip~\citep{sharpa2026}; our FPS for 10,000 taxels across 5 ContactDepthProbe sensors is $20\times$ Tacmap's and uses $7\times$ less GPU memory per environment.
    (e) HydroShear reports per-step time for a single $7\times 9$ (35 taxels) elastomer sensor on a robot arm; we rollout a trained robot-hand policy with five $5\times 4$ (100 taxels) elastomer sensors and achieve higher FPS as the number of parallel environments grows ( $1.6\times$ HydroShear's at 1024 envs).
    (f) Comparison against TacSL~\citep{akinola2025tacsl} reported FPS for their penalty-based force field vs. our KinematicTaxel sensor FPS and GPU memory usage for 10x10 and 100x100 taxels. We consistently achieve around 3x higher throughput (note the log scale on FPS) and can run at 16k envs without running out of memory.
    }
    \label{fig:prior_simulator_comparison}
\end{figure}

We integrate our tactile sensors into the open-source Genesis World physics simulator~\citep{genesis2026genesisworld}, exposing 7 sensor abstractions summarized in Table~\ref{tab:genesis_sensor_classes}, which together span the design space of current tactile hardware.
All sensors share a common pose-and-radius geometry, can be attached to arbitrary surfaces of any robot, and expose both a clean and a noisy readout under a configurable noise model (Appendix~\ref{app:sensor_noise_model}).
Our Elastomer sensor approach is based on HydroShear~\citep{dang_hydroshear_2026} and the Proximity sensor extends \citet{yin_learning_2025}; full equations are in Appendix~\ref{app:tactile_sensor_details}. In these prior works, these sensor simulation approaches have been validated to transfer to a real GelSight Mini on a robot gripper and ReSkin on the Allegro hand.
Our \texttt{ElastomerTaxel} extends HydroShear with an elastomer compressibility term and a clamped boundary condition, which lowers the marker-displacement error against real GelSight under both dilation and shear motion (Fig.~\ref{fig:elastomer_marker_comparison}).

Compared to our predecessors, \ours improves throughput by up to $20\times$ at matched sensor configurations and reduces GPU memory per environment by roughly $5\times$ (Fig.~\ref{fig:prior_simulator_comparison}), with the gap widening as the number of parallel environments grows.
Three implementation choices drive this.
First, the per-probe contact, depth, and force kernels are vectorized over probes \emph{and} environments, so a single launch covers the entire batch rather than iterating per sensor.
Second, mesh and point-cloud queries (SDF lookups, sphere--triangle intersection, proximity neighbor search) are accelerated by Bounding Volume Hierarchies (BVHs) over the tracked geometry, which keeps cost sublinear in the number of points.
Third, for the elastomer sensor's dilation kernel and the spatial crosstalk model, we exploit the regular planar taxel grid to replace the dense convolution by a 2D Fast Fourier Transform with separable kernels.
Together these allow us to scale \ours past $16{,}384$ parallel environments (Fig.~\ref{fig:method_sensor_benchmark}) and beyond 10,000 taxels per hand (Fig.~\ref{fig:prior_simulator_comparison}c) on a single GPU, which is the regime needed for student-policy training.

\subsection{Dexterous Task Training}

Our goal is to compare \emph{tactile observation types}, not specific hardware, on a fixed set of dexterous tasks.
We pair each simulator-level sensor class with a downstream postprocessing step to produce eight tactile observation types plus a proprioception-only baseline (\texttt{none}), listed in Table~\ref{tab:sensors}.
The \texttt{agg\_*} variants aggregate per-taxel signals to a single per-link value, matching the convention used by real fingertip force sensors in XHand1.
The per-taxel variants instead expose the full tactile field to the policy.
Holding placement, resolution, and policy architecture fixed across these types lets us isolate the effect of the abstraction itself.

\begin{table}[ht]
    \centering
    \footnotesize
    \begin{tabular}{@{}p{0.13\linewidth}p{0.16\linewidth}p{0.64\linewidth}@{}}
        \toprule
        Type & Sensor Type & Output (after optional postprocessing) \\
        \midrule
        \texttt{none} & -- & Proprioception-only baseline with tactile group removed. \\
        \texttt{bool} & ContactProbe & Binarized contact \textit{per taxel}. \\
        \texttt{agg\_bool} & ContactProbe & Binary contact \textit{per link} after thresholding by the number of taxels in contact. \\
        \texttt{depth} & ContactDepthProbe & Contact depth \textit{per taxel} \\
        \texttt{agg\_force} & ContactDepthProbe & Estimated force \textit{per link} based on summing contact depth along probe normals. \\
        \texttt{force} & KinematicTaxel & Force \textit{per taxel}. \\
        \texttt{force\_torque} & KinematicTaxel & Force and torque \textit{per taxel}. \\
        \texttt{elastomer} & ElastomerTaxel & XYZ marker displacement \textit{per taxel} (marker) on an elastomer surface. \\
        \texttt{proximity} & ProximityTaxel & Force and torque signals \textit{per taxel} based on  object surface proximity for signal strength. \\
        \bottomrule
    \end{tabular}
    \vspace{2pt}
    \caption{
    \textbf{Tactile observation types used by student policies.}
     We compare 8 different tactile representations, using the tactile signals from the simulated sensors and optionally postprocessing before passing into the observations.
     Tactile data \textit{per link} means we aggregate the data and output one tactile signal per every sensing area (e.g. 5 fingertips).
    }
    \label{tab:sensors}
\end{table}

\begin{figure}[ht]
    \centering
    \includegraphics[width=0.7\linewidth]{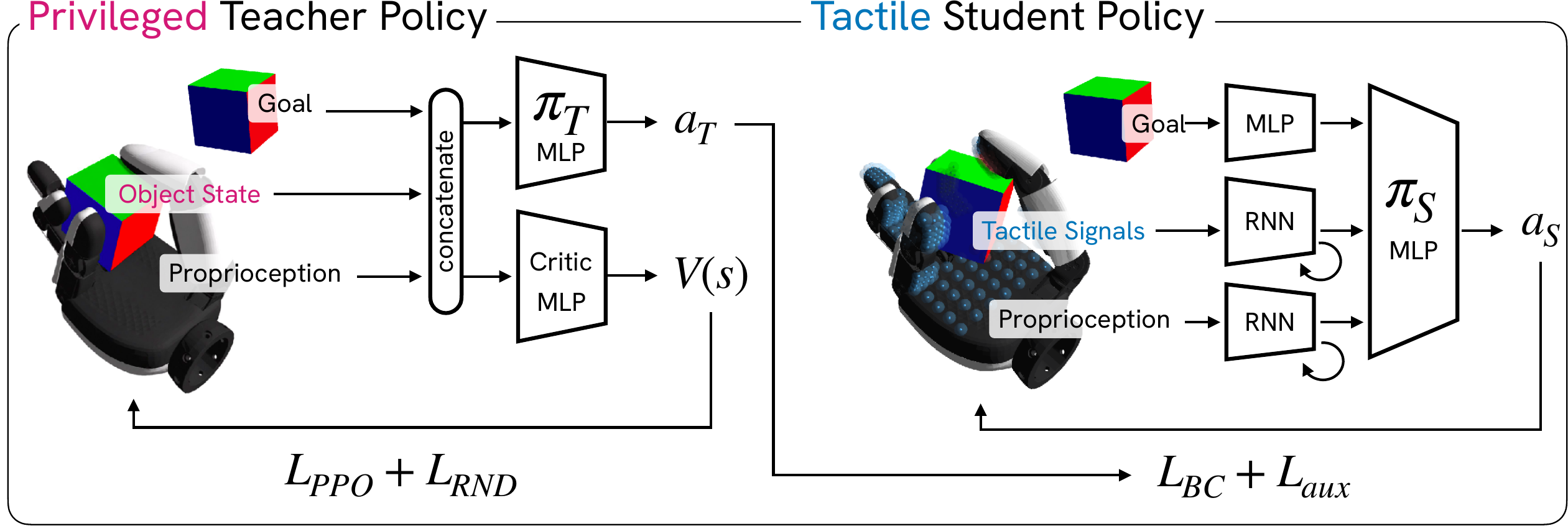}
    \caption{
    \textbf{Teacher-student training setup.}
    A privileged teacher is trained with PPO and an MLP actor-critic. We additionally incorporate a Random Network Distillation (RND)~\citep{burda_exploration_2018,schwarke_curiosity-driven_2023} loss to explore states more quickly.
    Tactile student policies encode each observation group before passing to the MLP head. In addition to the DAgger~\citep{dagger} behavioral cloning (BC) loss, we incorporate auxillary losses to decode object state. See Appendix \ref{app:training} for the full training parameters.}
    \label{fig:student_distillation}
\end{figure}

For each task--hand tuple, we first train a privileged teacher with PPO using full object state, then distill a tactile student that replaces the privileged state group with one of the tactile observation types from Table~\ref{tab:sensors} (Fig.~\ref{fig:student_distillation}).
The student is trained with behavioral cloning against the teacher's actions, plus auxiliary heads that decode privileged object state from the policy's hidden representation.
The decoders are not used at deployment; they act as a regularizer that pushes the tactile encoder to recover task-relevant object state from touch.
Each tactile observation group is processed by its own small MLP encoder before being concatenated with the proprioception features and fed to the policy head.
Full training hyperparameters and the per-task reward terms are listed in the Appendix.

We evaluate 3 tasks spanning complementary contact regimes:
\texttt{in\_palm\_rotate} requires reorienting an object on the palm with the thumb sweeping in to capture it, so locating the object before contact is informative.
\texttt{in\_hand\_repose} requires reposing an object to a target pose while it is in near-continuous contact with multiple fingers, so slip and grip-strength signals dominate.
\texttt{screwdriver} requires a fast finger gait that keeps a screwdriver spinning, so contacts are brief and rapidly changing.
We sweep three tactile placements (\texttt{tips}, \texttt{fingers}, \texttt{hand}) at three resolution levels, with both clean and noisy sensor settings; the full matrix is in Appendix Tables~\ref{tab:appendix_distill_matrix}--\ref{tab:appendix_noise_params}.

\subsection{Temperature Sensing for Learning Object Discrimination}
\label{sec:temperature_experiment}

\begin{figure}[ht]
    \centering
    \footnotesize
    \setlength{\tabcolsep}{6pt}
    \raisebox{-0.5\height}{\includegraphics[width=0.17\linewidth]{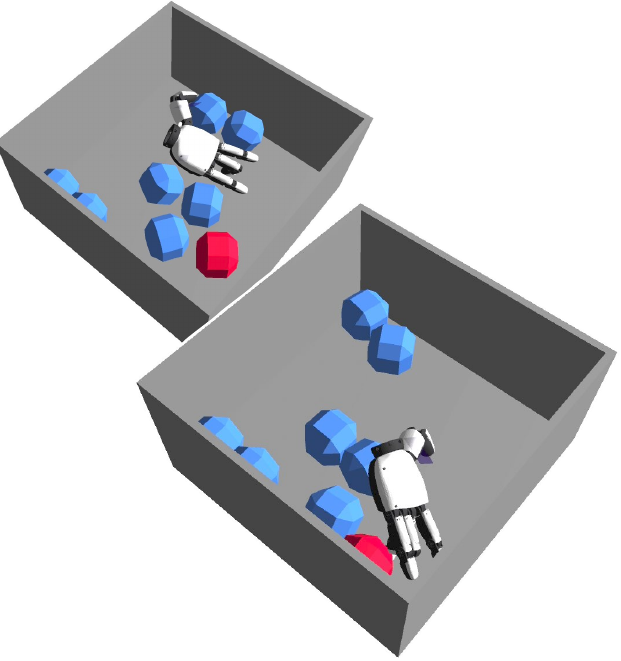}}%
    \hspace{1.5em}%
    \begin{tabular}[c]{@{}lcccc@{}}
        \toprule
        & \multicolumn{2}{c}{emissivity 0.5 (steel)} & \multicolumn{2}{c}{emissivity 0.85 (rubber)} \\
        \cmidrule(lr){2-3}\cmidrule(lr){4-5}
        conductivity & \multicolumn{4}{c}{heat generation (W/m$^2$)} \\
        \cmidrule(lr){2-5}
        (W/m$\cdot$K) & 500 & 5000 & 500 & 5000 \\
        \midrule
        1 (glass)      & --         & --         & -- & -- \\
        10 (steel)     & --         & --         & -- & -- \\
        100 (aluminum alloy) & --         & --         & -- & -- \\
        150 (aluminum) & \checkmark & \checkmark & -- & -- \\
        \bottomrule
    \end{tabular}
    \caption{
    \textbf{Temperature properties ablation for finding a target hot object.}
    Checkmark indicates that the hand successfully maintains touch with the hot ball.
    Emissivity values approximate stainless steel ($0.5$) and rubber ($0.85$); conductivities approximate glass ($1$), stainless steel ($10$), and aluminum ($100$).
    For scale, human skin dissipates on the order of $60\,$W/m$^2$ at rest and 100 to 600 W/m$^2$ during exercise, while small actuators under heavy load can reach $1{,}000$--$5{,}000\,$W/m$^2$.
    The temperature sensor implementation, task, and success metric are detailed in Appendix~\ref{app:temperature_experiment}.
    }
    \label{fig:temperature_experiment}
\end{figure}

Temperature can be a useful cue for telling objects apart, but temperature sensors on current robots are mostly used for hardware health monitoring rather than sensing, since the actuators heat up under sustained use.
Using our simulated tactile sensors, we explore what temperature sensitivity is needed to distinguish an object only by temperature.
We train a policy with proprioception and tactile sensors to find the hot ball out of 8 balls in a bin.
Our results in Fig.~\ref{fig:temperature_experiment} show that high sensitivity is indeed critical, and we fail to succeed on the task with material properties matching real temperature sensors on current robot hands.

\section{Results}
\label{sec:results}

\begin{figure}[ht]
    \centering
    \includegraphics[width=\linewidth]{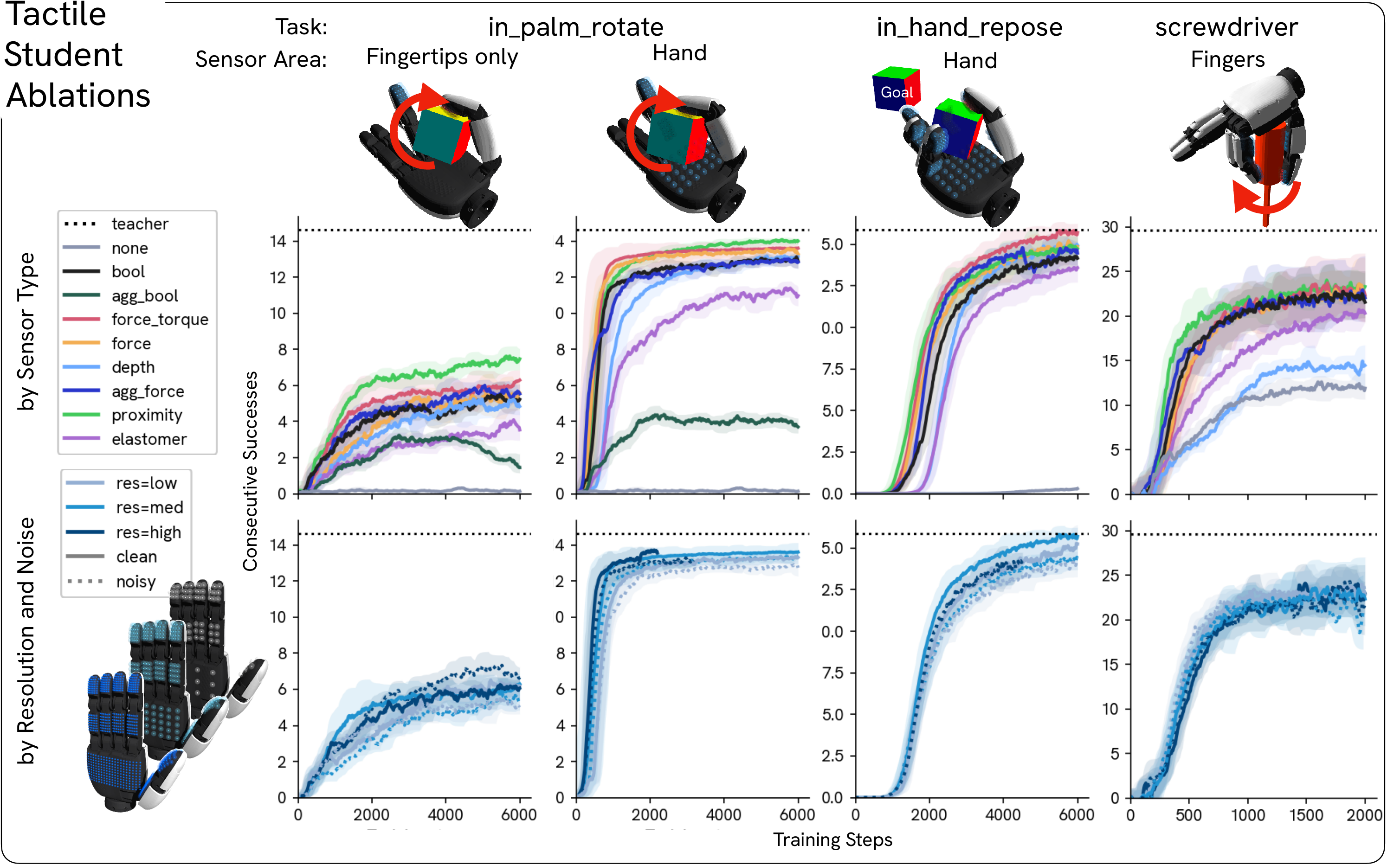}
    \caption{
    \textbf{Tactile student ablations.}
    For 3 tasks \texttt{in\_palm\_rotate}, \texttt{in\_hand\_repose}, and \texttt{screwdriver} using the XHand1, we compare tactile data types against the privileged teacher and distilled tactile student. For \texttt{in\_palm\_rotate} we try having fingertips only (the real XHand1 only has fingertips) vs including sensors on the whole hand. 
    We also vary the tactile resolution and add noise parameters (white noise, random walk drift, hysteresis, sensing radius noise) and compare.
    A complete table of parameters is available in the Appendix.
    }
    \label{fig:tactile_student_ablations}
\end{figure}

\subsection{Tactile Student Ablations}

\paragraph{Proprioception is not enough.}
In Fig~\ref{fig:tactile_student_ablations}, we can see that the \texttt{none} baseline trails every tactile student on all three tasks, including the cheapest binary contact variant.
The auxiliary state decoders alone are not enough to recover task-relevant object state from proprioception; any meaningful student performance requires the tactile group.

\paragraph{Placement dominates sensor type.}
Restricting sensing to fingertips, the placement that most current commercial hardware supports, trails whole-hand coverage by a large margin on \texttt{in\_palm\_rotate}.
Adding the palm and mid-finger surfaces closes most of the remaining gap to the privileged teacher, even for sensor types that individually carry less information.
Adding taxels on the palm and proximal phalanges of the fingers is, at the margin, more useful than upgrading the fingertip sensor.

\paragraph{The best sensor type is task-dependent, with force/torque as a robust default.}
On \texttt{in\_hand\_repose}, where the object is in near-continuous contact and the dominant failure mode is incipient slip, the \texttt{force\_torque} student is best and clearly separates from the binary and depth variants.
On \texttt{in\_palm\_rotate}, \texttt{proximity} edges out the contact-only types, since its sensing radius registers the approaching object before direct contact and lets the thumb pre-shape rather than search.
On \texttt{screwdriver}, where contacts are brief and the fingers gait quickly, all tactile signals perform similarly and none saturates the teacher; we conjecture that integration over time, or visual feedback, is the missing channel here rather than a different contact abstraction.
Aggregated across tasks, per-taxel \texttt{force\_torque} matches or outperforms every other type and is our recommended default when hardware allows.

\paragraph{Elastomer displacement underperforms when per-taxel locality matters.}
The \texttt{elastomer} student trails \texttt{force\_torque} on both \texttt{in\_palm\_rotate} and \texttt{in\_hand\_repose}.
This is consistent with how the substrate model behaves: marker displacement at one taxel is a function of indentation \emph{and} shear at neighboring taxels, so a displacement in $x$ can be caused by adjacent dilation as easily as by local shear.
The resulting signal is well suited for inferring object shape patches, the use case GelSight-like sensors were originally designed for, but less suited for reading off the local force vector that the in-hand tasks actually need.

\paragraph{Sim-to-real validation.}
We validate transfer by deploying the \texttt{in\_palm\_rotate} policy on the real XHand1, whose only tactile sensing is fingertip aggregate force.
The deployed policy achieves one to two successes, which matches the success rate of the fingertip \texttt{agg\_bool} student in simulation, the observation type that most closely mirrors the real fingertip readout.
This confirms that policies trained in \ours transfer to hardware, and that the simulated fingertip abstraction is a faithful enough proxy for the real sensor to predict its performance.

\section{Discussion and Limitations}
Across all tasks, sensor types, and placements, the takeaways for practitioners outfitting a dexterous hand are threefold.
First, cover the palm and proximal phalanges before paying for higher-end fingertip sensors.
Second, prefer per-taxel force/torque as a default abstraction.
Third, consider a proximity channel when the task involves capturing an object that approaches the hand rather than one already grasped.
The first two findings are at odds with the de facto convention in current commercial tactile fingertips, which concentrate spatial resolution on the distal pad and stop there.
From a simulation-design perspective, our experiments also suggest that the dominant source of useful tactile information for these tasks is the coarse spatial distribution of contact rather than the fine-grained mechanics of the substrate.
This is encouraging for rigid-body tactile simulation, since substrate physics is exactly what is most expensive to model faithfully.

We deliberately scope this paper to sensor implementation and observation-type comparison.
Because our students distill from a privileged teacher, they inherit its strategy and are therefore bounded by it.
An important direction for future work would be a larger sweep over hands and tasks, with continued (curiosity-driven) RL~\citep{burda_exploration_2018} using either tactile observations as the only sensory channel or paired with vision, would help separate ``what the teacher can be matched on'' from ``what touch is actually capable of.''

\section*{Acknowledgements}
This material is based upon work supported by the National Science Foundation Graduate Research Fellowship Program under Grant No(s) DGE2140739. Any opinions, findings, and conclusions or recommendations expressed in this material are those of the author(s) and do not necessarily reflect the views of the National Science Foundation.
This work is also partially funded by DARPA SAFRON award HR0011-25-3-0203 and an Amazon robotics award.
A.N. thanks the Burroughs Wellcome Fund (CASI award) and Google Robotics Award for funding.

\clearpage

\bibliography{references}

\clearpage
\appendix

\section{Tactile Sensor Implementation}
\label{app:tactile_sensor_details}

This section gives the mathematical definitions of our tactile sensor implementations, which have been incrementally integrated into Genesis World open-source physics simulation platform.
All code used for this work will be made available on GitHub.
Note that there may be some differences in the latest version of Genesis compared to the version used in this paper; please refer to the code for the most accurate source of truth.

\subsection{Probe geometry and contact-depth query}

Each probe is rigidly attached to a sensor link and described by a link-frame position, unit normal, and radius.
Let sensor link $L$ have pose $(p_L,R_L)$ and let probe $j$ have link-frame position $x_j$, unit normal $n_j$, and radius $R_j$.
Its world-frame position and normal are
\begin{equation}
    q_j=p_L+R_Lx_j,
    \qquad
    a_j=R_Ln_j.
\end{equation}
Depth and force probes only consider collision geometries that are currently in a rigid-body contact pair with the sensor link.
A probe with $R_j=0$ is treated as an inactive filler and returns zero, which keeps batched tensors regular over heterogeneous link layouts.

\paragraph{Two contact-depth backends.}
The penetration depth at probe $j$ can be queried in two ways, selected by the \texttt{contact\_depth\_query} option.
The \texttt{sdf} backend queries the per-geometry analytic signed-distance field maintained by the rigid solver.
For an opposing collision geometry $g$ with SDF $\phi_g$, the penetration depth is
\begin{equation}
    d^{\mathrm{sdf}}_j
    =
    \max_g [R_j-\phi_g(q_j)]_+,
\end{equation}
and the contact normal is the SDF gradient $m_j=\nabla \phi_{g^\star}(q_j)$ at the winning geometry $g^\star$ when $d^{\mathrm{sdf}}_j>0$.
This is fast and exact on primitives, but requires SDF activation on the collider.
The \texttt{raycast} backend instead walks a per-frame Bounding Volume Hierarchy over the rigid-body collision meshes, shared with Genesis's raycaster sensor.
At each candidate BVH leaf, the probe runs a sphere--triangle closest-point test (penetration $=R_j-\mathrm{dist}$) and a ray--triangle test along $-a_j$ (penetration $=R_j-\mathrm{hit\_distance}$); the deepest of the two wins, and the contact normal is the face normal of the deepest-penetrating triangle.
The two backends produce equivalent depths to leading order for smooth meshes and differ at sharp features.
\texttt{sdf} is the default; \texttt{raycast} handles arbitrary triangle meshes uniformly and is preferable when the scene mixes many small or thin objects.

\subsection{\texttt{SurfaceDistanceProbe}}

The \texttt{SurfaceDistanceProbe} reports, per probe, the shortest distance from the probe center to the surface of each tracked object.
Unlike the contact sensors below, it does not require a rigid-body contact pair and is defined before contact, which makes it a pre-contact proximity cue.

\subsection{\texttt{ContactProbe}}

The \texttt{ContactProbe} binarizes the measured penetration depth $d^{\mathrm{m}}_{bj}$ at probe $j$ into a contact bit, with optional Schmitt hysteresis to suppress chatter near the contact boundary:
\begin{equation}
    y^{\mathrm{contact}}_{bj}(t)
    =
    \mathbb{I}\left[
    d^{\mathrm{m}}_{bj}(t)>\eta_{\mathrm{on}}
    \ \lor\
    \left(y^{\mathrm{contact}}_{bj}(t-1)=1
    \land
    d^{\mathrm{m}}_{bj}(t)>\eta_{\mathrm{off}}\right)
    \right].
\end{equation}
Here $\eta_{\mathrm{on}}$ is the contact threshold and $\eta_{\mathrm{off}}\le\eta_{\mathrm{on}}$ the release threshold; setting $\eta_{\mathrm{off}}=\eta_{\mathrm{on}}$ disables the hysteresis.
We use $\eta_{\mathrm{on}}=5\times10^{-4}$ m, adding a release threshold $\eta_{\mathrm{off}}=2\times10^{-4}$ m and dead-taxel noise in the noisy condition (Table~\ref{tab:appendix_noise_params}).

\subsection{\texttt{ContactDepthProbe}}

The \texttt{ContactDepthProbe} reports the raw measured penetration depth $d^{\mathrm{m}}_{bj}$ at each probe, taken from the \texttt{sdf} or \texttt{raycast} backend of the previous section.

\subsection{\texttt{KinematicTaxel}}

The \texttt{KinematicTaxel} converts depth, contact normal, and relative velocity into a per-taxel six-channel force/torque estimate without modeling a deformable substrate.
Let $\bar m_j=R_L^\top m_j$ be the SDF normal in the sensor-link frame.
For a contact with opposing link $C$, the point velocities are
\begin{align}
    v_C(q_j) &= v_C+\omega_C\times(q_j-r_C),\\
    v_L(q_j) &= v_L+\omega_L\times(q_j-r_L),
\end{align}
where $r_C,r_L$ are the corresponding centers of mass. The relative velocity in the sensor-link frame is
\begin{equation}
    v_{\mathrm{rel},j}
    =
    R_L^\top\left(v_C(q_j)-v_L(q_j)\right).
\end{equation}
With $s_j=(d_j)^\alpha$, the normal and tangential components are
\begin{equation}
    v_{n,j}=(v_{\mathrm{rel},j}^{\top}\bar m_j)\bar m_j,
    \qquad
    v_{t,j}=v_{\mathrm{rel},j}-v_{n,j}.
\end{equation}
The local force and torque are
\begin{align}
    f_j
    &=
    k_n s_j \bar m_j
    + c_n s_j v_{n,j}
    - k_t v_{t,j},\\
    \tau_j
    &=
    x_j\times f_j
    -
    k_\omega
    \left((\omega_C-\omega_L)^\top m_j\right)\bar m_j.
\end{align}
If no opposing contact link is available we drop the velocity terms, returning only the spring force and $\tau_j=x_j\times f_j$.
The sensor reports the six channels $(f_j,\tau_j)$ per taxel.
The parameters used in our experiments are
\begin{equation}
    k_n=500,\qquad c_n=1,\qquad \alpha=1.2,\qquad k_t=2,\qquad k_\omega=2.
\end{equation}

For regular planar taxel grids, we apply spatial crosstalk to the noisy channel.
Let $z_c$ be one force or torque channel on the grid, $\chi$ the crosstalk strength, and $G_\sigma$ an L1-normalized Gaussian kernel of standard deviation $\sigma$.
The channel update is
\begin{equation}
    \tilde z_c
    =
    (1-\chi)z_c+\chi(G_\sigma*z_c),
\end{equation}
applied independently to all six channels.

\subsection{\texttt{ProximityTaxel}}

The \texttt{ProximityTaxel} models a geometry-aware taxel that responds to nearby tracked mass, mirroring how capacitive and magnetic skins behave.
Tracked links are sampled into a surface point cloud at scene reset and organized in a static BVH for fast neighbor queries.
For taxel $j$, let $\mathcal{P}_j=\{i:\lVert p_i-q_j\rVert<R_j\}$ be the set of tracked points inside the taxel sensing sphere.
Each point contributes
\begin{equation}
    P_{ij}=R_j-\lVert p_i-q_j\rVert.
\end{equation}
Let
\begin{equation}
    v_{\mathrm{tax},j}=v_L+\omega_L\times(q_j-r_L),
    \qquad
    v_{p_i}=v_i+\omega_i\times(p_i-r_i),
\end{equation}
and remove the taxel-normal component of relative velocity:
\begin{equation}
    v_{t,ij}
    =
    (v_{p_i}-v_{\mathrm{tax},j})
    -
    a_j
    \left((v_{p_i}-v_{\mathrm{tax},j})^\top a_j\right).
\end{equation}
The density scale is
\begin{equation}
    \rho_b=\frac{\texttt{density\_scalar}}{\max(N_{\mathrm{active},b},1)}.
\end{equation}
The world-frame force and torque are
\begin{align}
    f^w_{bj}
    &=
    k\rho_b\sum_{i\in\mathcal{P}_j}P_{ij}a_j
    +
    k_s\rho_b\sum_{i\in\mathcal{P}_j}P_{ij}v_{t,ij},\\
    \tau^w_{bj}
    &=
    k\rho_b\sum_{i\in\mathcal{P}_j}P_{ij}
    \left((p_i-q_j)\times a_j\right).
\end{align}
The reported channels are $R_L^\top f^w_{bj}$ and $R_L^\top\tau^w_{bj}$.
Our experiments use $R_j=0.01$ m, $N_{\mathrm{pc}}=5000$, $k=300$, $k_s=10$, and a density scalar of 100.
Radius noise enters by perturbing $R_j$ before the BVH query; per-probe gain scales the accumulated penetration, slip, and torque terms before the force/torque conversion.

\subsection{\texttt{ElastomerTaxel}}

The \texttt{ElastomerTaxel} models marker displacements on a deformable substrate by combining SDF-driven dilation with anchored shear history~\citep{dang_hydroshear_2026}.
Let
\begin{equation}
    \Pi_n(z)=z-(z^\top n)n
\end{equation}
be projection onto the tangent plane.
For source probe $i$, the per-probe penetration into the elastomer is
\begin{equation}
    h_i=\max\left(0,-\min_g\phi_g(q_i)\right).
\end{equation}
For target marker $j$, write $r_{ij}=\Pi_{n_j}(x_j-x_i)$ for the tangential offset to source $i$.
The in-plane marker spreading interpolates between a local Gaussian falloff and an incompressible global stretch through a compressibility parameter $c\in[0,1]$:
\begin{equation}
    w_{ij}
    =
    h_i\left(
        c\,\frac{g(r_{ij})}{\bar g}
        +
        (1-c)\,\frac{\rho(r_{ij})}{\bar\rho}
    \right),
\end{equation}
where $g(r)=\exp(-\lambda_d\lVert r\rVert^2)$ is the Gaussian kernel, $\rho(r)=(\lVert r\rVert^2+\varepsilon^2)^{-1}$ is a regularized inverse-distance kernel, and $\bar g=e^{-1/2}/\sqrt{2\lambda_d}$ and $\bar\rho=1/(2\varepsilon)$ peak-normalize the two so that $c$ blends them on a common scale.
Setting $c=1$ recovers the local Gaussian response, while $c=0$ recovers an incompressible response whose in-plane displacement decays as $1/\lVert r\rVert$.
The out-of-plane bulge keeps the Gaussian falloff with the depth power law, and the dilation contribution is
\begin{equation}
    u^{d}_{ij}
    =
    s_d\left(
        r_{ij}\,w_{ij}
        +
        n_j\,h_i^{\alpha}\,g(r_{ij})
    \right).
\end{equation}
The regularization length $\varepsilon$ is set to the elastomer thickness when it is modeled, otherwise to the taxel spacing.
Tracked links are also sampled into surface points to capture shear history.
A sampled point $p$ initializes an anchor $e_p$ when its elastomer SDF drops below $-\eta_{\mathrm{enter}}$ and clears the anchor when the SDF rises above $+\eta_{\mathrm{exit}}$.
Let $z_p$ be the current sensor-frame point position and $h_p$ the elastomer penetration depth.
The shear contribution to target marker $j$ is
\begin{equation}
    u^{s}_{pj}
    =
    s_s
    \Pi_{n_j}(z_p-e_p)
    h_p
    \exp\left(
        -\lambda_s
        \left\lVert\Pi_{n_j}(x_j-z_p)\right\rVert^2
    \right).
\end{equation}
The displacement output is
\begin{equation}
    y^{\mathrm{elastomer}}_j
    =
    \sum_i u^d_{ij}+\sum_p u^s_{pj}.
\end{equation}

\paragraph{Clamped boundary condition.}
When the elastomer is bonded to a rigid housing, markers cannot move at the pad edge.
We optionally enforce this as a Dirichlet condition on the in-plane dilation field.
Writing the field above as $u^{\mathrm{free}}$, the corrected field is $u=u^{\mathrm{free}}+u^{\mathrm{corr}}$, where the correction solves the homogeneous depth-averaged bonded-layer equations
\begin{equation}
    -\nabla^2 u^{\mathrm{corr}}
    +
    \frac{\beta}{t^2}\,u^{\mathrm{corr}}
    -
    K\,\nabla(\nabla\cdot u^{\mathrm{corr}})
    =
    0
\end{equation}
on the pad interior with $u^{\mathrm{corr}}=-u^{\mathrm{free}}$ on the boundary loop, so that the total displacement vanishes at the wall over a boundary layer of width $t/\sqrt{\beta}$.
Here $t$ is the elastomer thickness, $\beta=3$ sets the drag, and $K$ penalizes in-plane compression.
On the regular grid we precompute this discrete solution operator once at build time and apply it as a low-rank correction to the FFT field at each step.

Our experiments use $N_{\mathrm{pc}}=1000$, $s_d=100$, $s_s=100$, $\lambda_d=8000$, $\lambda_s=2000$, $\alpha=1.2$, compressibility $c=1$ (the local-Gaussian limit), and shear enter/exit thresholds $\eta_{\mathrm{enter}}=10^{-5}$ m and $\eta_{\mathrm{exit}}=10^{-4}$ m, with no clamped boundary.
The compressibility and boundary condition are calibrated against real GelSight marker motion in Fig.~\ref{fig:elastomer_marker_comparison}.
For regular planar grids we accelerate the dilation term with FFT: tangent channels convolve $h_i$ and the normal channel convolves $h_i^\alpha$; shear is accumulated directly.

\subsection{Sensor Noise Model}
\label{app:sensor_noise_model}

Every sensor in Genesis exposes both a clean ground-truth readout and a noisy readout.
Table~\ref{tab:appendix_noise_capabilities} lists which imperfection knobs each sensor type supports; the numerical values we use in the experiments are in Section~\ref{app:training} (Table~\ref{tab:appendix_noise_params}).

\paragraph{Base noise (every sensor).}
At the simulator level, every sensor honors a read delay $\Delta_{\mathrm{read}}$ (with optional uniform jitter), additive Gaussian white noise $\sigma$, constant bias $b$, random-walk drift $\sigma_{\mathrm{rw}}$, and uniform quantization with step $q$:
\begin{equation}
    \tilde y_t = \operatorname{quantize}\big(y_{t-\Delta_{\mathrm{read}}} + b + n_t + d_t,\, q\big),
    \quad
    n_t\sim\mathcal{N}(0,\sigma^2),
    \quad
    d_t = d_{t-1} + w_t,\ w_t\sim\mathcal{N}(0,\sigma_{\mathrm{rw}}^2).
\end{equation}

\paragraph{Probe-level imperfections (probe-based sensors).}
Probe sensors additionally perturb the per-probe sensing radius and apply a per-probe multiplicative gain:
\begin{equation}
    \tilde R_j = \max(0,\, R_j + \epsilon_j),
    \qquad
    \epsilon_j\sim\mathcal{U}[-\rho_j,\rho_j],
\end{equation}
\begin{equation}
    \gamma_{bj}\sim\mathcal{U}[\gamma_{\min,j},\gamma_{\max,j}]\ \text{at reset},
\end{equation}
where the gain $\gamma_{bj}$ models persistent per-unit calibration error.

\paragraph{Taxel-level imperfections (tactile probe sensors).}
Tactile probe sensors further sample dead taxels at episode reset:
\begin{equation}
    M_{bj}\sim\operatorname{Bernoulli}(p_j),
    \qquad
    z_{bj}\sim\mathcal{U}[\ell_j,u_j],
\end{equation}
and when $M_{bj}=1$ the measured value at taxel $j$ is overwritten by the stuck readout $z_{bj}$ for the entire episode.

\paragraph{Viscoelastic hysteresis (contact, depth, kinematic, elastomer, proximity).}
Substrate-like sensors model a single-Maxwell viscoelastic loop on the noisy readout:
\begin{align}
    \xi_t &= \exp(-\Delta t/\tau)\,\xi_{t-1} + (x_t-x_{t-1}),\\
    \tilde x_t &= x_t + \beta\,\xi_t,
\end{align}
with strength $\beta$ and time constant $\tau$.
After a rising step the noisy signal overshoots and decays back to equilibrium; after a falling step it undershoots analogously.

\paragraph{Spatial crosstalk.}
On regular planar grids, the sensor mixes neighboring channels to model spatial crosstalk between adjacent sensing sites.
Let $z_c$ be one force or torque channel on the grid, $\chi$ the crosstalk strength, and $G_\sigma$ an L1-normalized Gaussian kernel of standard deviation $\sigma$:
\begin{equation}
    \tilde z_c = (1-\chi)z_c + \chi(G_\sigma * z_c),
\end{equation}
applied independently to each of the six channels.

\begin{table}[htbp]
    \centering
    \small
    \setlength{\tabcolsep}{3pt}
    \begin{tabular}{@{}lcccccc@{}}
        \toprule
        Noise knob & SurfaceDist. & ContactProbe & ContactDepth & Kinematic & Elastomer & Proximity \\
        \midrule
        Read delay / jitter & \checkmark & \checkmark & \checkmark & \checkmark & \checkmark & \checkmark \\
        White noise $\sigma$ & \checkmark & \checkmark & \checkmark & \checkmark & \checkmark & \checkmark \\
        Constant bias       & \checkmark & \checkmark & \checkmark & \checkmark & \checkmark & \checkmark \\
        Random-walk drift   & \checkmark & \checkmark & \checkmark & \checkmark & \checkmark & \checkmark \\
        Quantization        & \checkmark & \checkmark & \checkmark & \checkmark & \checkmark & \checkmark \\
        Probe-radius noise  & \checkmark & \checkmark & \checkmark & \checkmark & \checkmark & \checkmark \\
        Per-probe gain      & --         & \checkmark & \checkmark & \checkmark & \checkmark & \checkmark \\
        Dead taxels         & --         & \checkmark & \checkmark & \checkmark & \checkmark & \checkmark \\
        Viscoelastic hysteresis & --     & \checkmark & \checkmark & \checkmark & \checkmark & \checkmark \\
        Spatial crosstalk   & --         & --         & --         & \checkmark & --         & \checkmark \\
        \bottomrule
    \end{tabular}
    \vspace{2pt}
    \caption{Imperfection knobs available per sensor type.}
    \label{tab:appendix_noise_capabilities}
\end{table}

\clearpage
\section{Training Setup}
\label{app:training}

The training pipeline, task definitions, and tactile sensor wiring used in this paper live in the companion \texttt{dexterous\_hands} repository.
This section documents the optimization and network hyperparameters, the observation--type mapping, the dexterous tasks, the sweep matrix, the per-resolution probe counts, and the noise values used in the noisy condition.

\subsection{Optimization and Network Hyperparameters}

Table~\ref{tab:appendix_hparams} lists the teacher PPO, RND exploration, and student DAgger optimization settings; values that differ across tasks are shown per task.
Teacher and student share the same actor--critic backbone: a three-layer MLP head with hidden sizes $[512,256,128]$ and ELU activations, with input normalization and an initial policy standard deviation of $1.0$.
Each non-tactile observation group is embedded by a per-group encoder (an MLP $[512,256,128]\!\to\!64$, or an LSTM with hidden size $128\!\to\!64$) before concatenation.
Each tactile group is embedded by its own encoder projecting to a $32$-dimensional embedding: gridless types use an MLP ($[64,64]$) or an LSTM (hidden $64$), while grid-structured types use a convolutional encoder (\texttt{tactile\_cnn}, channels $[16,32]$, kernel $3$) or a convolutional--recurrent encoder (\texttt{tactile\_convrnn}, $16$ channels, kernel $3$, layer norm).
The RND predictor uses learning rate $10^{-3}$, an $8$-dimensional output, and normalized state and reward.
The student additionally trains auxiliary heads that decode privileged object state from its latent (Table~\ref{tab:appendix_aux_losses}); these decoders are used only during training and discarded at deployment.
The simulation steps at $200$ Hz with a control decimation of $5$ ($40$ Hz control).

\begin{table}[htbp]
    \centering
    \small
    \begin{tabular}{@{}lccc@{}}
        \toprule
        & \texttt{in\_palm\_rotate} & \texttt{in\_hand\_repose} & \texttt{screwdriver} \\
        \midrule
        \multicolumn{4}{@{}l}{\textit{Teacher (PPO)}}\\
        \quad parallel environments & 8192 & 8192 & 8192 \\
        \quad steps per env & 24 & 24 & 12 \\
        \quad teacher iterations & 6000 & 20000 & 5000 \\
        \quad learning rate (adaptive) & $10^{-3}$ & $10^{-3}$ & $10^{-3}$ \\
        \quad learning epochs / minibatches & 5 / 4 & 5 / 4 & 5 / 4 \\
        \quad discount $\gamma$ & 0.998 & 0.998 & 0.99 \\
        \quad GAE $\lambda$ & 0.95 & 0.95 & 0.95 \\
        \quad clip param / desired KL & 0.2 / 0.01 & 0.2 / 0.01 & 0.2 / 0.01 \\
        \quad entropy coef. & $2\times10^{-3}$ & $2\times10^{-3}$ & 0 \\
        \quad value loss coef. / max grad norm & 1.0 / 1.0 & 1.0 / 1.0 & 1.0 / 1.0 \\
        \quad episode length (s) & 10 & 20 & 20 \\
        \midrule
        \multicolumn{4}{@{}l}{\textit{Student (DAgger)}}\\
        \quad student iterations & 6000 & 6000 & 2000 \\
        \quad learning rate & $10^{-4}$ & $10^{-4}$ & $10^{-5}$ \\
        \quad learning epochs & 5 & 5 & 5 \\
        \quad gradient length & 2 & 2 & 4 \\
        \quad clip param & -- & -- & 0.1 \\
        \quad max grad norm & 2.0 & -- & 0.5 \\
        \quad BC loss & MSE & MSE & inv-var MSE \\
        \bottomrule
    \end{tabular}
    \vspace{2pt}
    \caption{Teacher PPO and student DAgger optimization hyperparameters.}
    \label{tab:appendix_hparams}
\end{table}

\begin{table}[htbp]
    \centering
    \small
    \begin{tabular}{@{}llccc@{}}
        \toprule
        Task & Decoded target & Weight & Target scale & Head hidden dims \\
        \midrule
        \texttt{in\_palm\_rotate} & object size & 1.0 & 20.0 & $[128,64]$ \\
        \texttt{in\_palm\_rotate} & goal distance & 1.0 & 1.0 & $[128,64]$ \\
        \texttt{screwdriver} & object tilt & 0.5 & 1.0 & $[512,128]$ \\
        \texttt{screwdriver} & rotation progress & 0.5 & 1.0 & $[512,128]$ \\
        \texttt{in\_hand\_repose} & \multicolumn{4}{l}{none} \\
        \bottomrule
    \end{tabular}
    \vspace{2pt}
    \caption{Auxiliary decoder losses used during student distillation.
    Each head predicts a privileged scalar from the student latent under an MSE loss (target scaled by the listed factor); the total loss adds the weighted auxiliary terms to the DAgger behavioral-cloning loss.
    The decoders regularize the tactile encoder toward task-relevant object state and are discarded at deployment.}
    \label{tab:appendix_aux_losses}
\end{table}

\subsection{Tactile Observation Types}

Table~\ref{tab:appendix_sensor_registry} shows how each downstream observation type maps to an underlying Genesis sensor abstraction together with the postprocessing applied to the raw read.

\begin{table}[htbp]
    \centering
    \small
    \begin{tabular}{@{}p{0.14\linewidth}p{0.22\linewidth}p{0.30\linewidth}p{0.22\linewidth}@{}}
        \toprule
        Type & Sensor abstraction & Key parameters & Policy observation \\
        \midrule
        \texttt{bool} & \texttt{ContactProbe} & $\eta_{\mathrm{on}}=5\times10^{-4}$ m & thresholded contact bits \\
        \texttt{agg\_bool} & \texttt{ContactProbe} & $\eta_{\mathrm{on}}=5\times10^{-4}$ m; count threshold $>2$ & per-link aggregate contact bit \\
        \texttt{depth} & \texttt{ContactDepthProbe} & \texttt{contact\_depth\_query}\,$=$\,\texttt{sdf} & flattened contact depths \\
        \texttt{agg\_force} & \texttt{ContactDepthProbe} & \texttt{sdf} query; $10^4$ scale, ZYX axis order & per-link patch force \\
        \texttt{force} & \texttt{KinematicTaxel} & $k_n=500$, $c_n=1$, $\alpha=1.2$, $k_t=2$, $k_\omega=2$ & force channels only \\
        \texttt{force\_torque} & \texttt{KinematicTaxel} & same as \texttt{force} & force and torque channels \\
        \texttt{elastomer} & \texttt{ElastomerTaxel} & $N_{\mathrm{pc}}=1000$, $\lambda_d=8000$, $\lambda_s=2000$, $s_d=100$, $s_s=100$, $\alpha=1.2$ & marker displacements \\
        \texttt{proximity} & \texttt{ProximityTaxel} & $R=0.01$ m, $N_{\mathrm{pc}}=5000$, $k=300$, $k_s=10$ & force and torque channels \\
        \bottomrule
    \end{tabular}
    \vspace{2pt}
    \caption{Mapping from downstream observation type to underlying Genesis sensor abstraction and key parameters.}
    \label{tab:appendix_sensor_registry}
\end{table}

Each observation type derives from one native sensor (Appendix~\ref{app:tactile_sensor_details}) followed by light postprocessing.
Writing $d^{\mathrm{m}}_{bj}$ for the measured \texttt{ContactDepthProbe} depth and $n_j$ for the link-frame probe normal:

\paragraph{Contact observations.}
\texttt{depth} flattens the per-taxel \texttt{ContactDepthProbe} depths.
\texttt{bool} is the per-taxel \texttt{ContactProbe} bit, and \texttt{agg\_bool} sums those bits over a link and reports a single bit when more than two taxels are in contact.
\texttt{agg\_force} aggregates the contact-depth probes into one patch force per link,
\begin{equation}
    y^{\mathrm{agg\_force}}_b
    =
    10^4\, P_{\mathrm{ZYX}}\sum_j d^{\mathrm{m}}_{bj}\, n_j,
    \qquad
    P_{\mathrm{ZYX}}(f_x,f_y,f_z)=(f_z,f_y,f_x),
\end{equation}
permuting axes and applying an empirical scale to match the per-fingertip force that the XHand1 \texttt{calc\_pressure} channel returns on the real robot.

\paragraph{Force, displacement, and proximity observations.}
\texttt{force} and \texttt{force\_torque} read the \texttt{KinematicTaxel}: \texttt{force\_torque} keeps all six channels $(f_j,\tau_j)$, while \texttt{force} drops the torque, $h_{\mathrm{force}}(F,T)=\operatorname{vec}(F)$.
\texttt{proximity} flattens the \texttt{ProximityTaxel} force and torque channels, and \texttt{elastomer} flattens the \texttt{ElastomerTaxel} marker displacements.

Except for \texttt{agg\_bool}, \texttt{agg\_force}, and \texttt{force}, the policy input flattens each raw sensor tensor and concatenates the results:
\begin{equation}
    h_{\mathrm{flat}}(Y_1,\ldots,Y_K)
    =
    \operatorname{concat}\big(
    \operatorname{vec}(Y_1),\ldots,\operatorname{vec}(Y_K)
    \big).
\end{equation}

Only the per-link aggregate types \texttt{agg\_bool} and \texttt{agg\_force} have a counterpart on the real XHand1; their deployment is described in Appendix~\ref{app:sim_to_real}.

\subsection{Tasks and Hands}

\textbf{\texttt{in\_palm\_rotate}.}
A partial-hand task in which the policy controls only the thumb and middle finger (index, ring, and pinky fingers are frozen) to rotate the object around the commanded axis. the object is reset to a sampled bottom-aligned pose at the center of the palm.
We train with 16 objects: 14 different cubes and 2 cylinders around 4cm in length.

\textbf{\texttt{in\_hand\_repose}.}
A whole-hand reposing task in which the policy must drive the object pose to a target orientation while keeping it in continuous contact with multiple fingers.
The reward combines orientation tracking with grip-strength and slip penalties.
We train with the same 16 objects as in the previous task.

\textbf{\texttt{screwdriver}.}
A finger-gaiting task in which the hand spins a screwdriver about its long axis while keeping the tip engaged.
We train with 4 different screwdriver handles.

\paragraph{Observations.}
All three tasks share the same observation grouping (Table~\ref{tab:appendix_obs_groups}).
The student (deployable) policy sees only proprioception (joint positions and velocities and the last action), the tactile group (one observation type from Table~\ref{tab:sensors}), and, for the reorientation tasks, the goal.
The privileged groups available to the teacher and critic add per-link contact forces, full object state, and object properties.
\texttt{screwdriver} has no goal group and folds the surface-distance probes into its object-state group.

\begin{table}[htbp]
    \centering
    \small
    \begin{tabular}{@{}lll@{}}
        \toprule
        Group & Contents & Available to \\
        \midrule
        \texttt{proprio} & joint positions, joint velocities, last action & student + teacher \\
        \texttt{tactile\_sensors} & one tactile observation type (Table~\ref{tab:sensors}) & student \\
        \texttt{goal} & goal orientation and angular error to target & student\,$^\ddagger$ \\
        \texttt{priv\_proprio} & proprioception plus per-link contact forces & teacher, critic \\
        \texttt{priv\_obj\_state} & object position, 6D orientation, linear and angular velocity & teacher, critic \\
        \texttt{priv\_obj\_props} & object friction, mass, and surface-distance probes & critic \\
        \bottomrule
    \end{tabular}
    \vspace{2pt}
    \caption{Observation groups. The student policy is restricted to \texttt{proprio}, \texttt{tactile\_sensors}, and (where present) \texttt{goal}; the privileged groups are used only to train the teacher and critic.
    $^\ddagger$\texttt{in\_palm\_rotate} and \texttt{in\_hand\_repose} only; \texttt{screwdriver} has no goal group and places the surface-distance probes in \texttt{priv\_obj\_state}.}
    \label{tab:appendix_obs_groups}
\end{table}

\paragraph{Reward terms.}
Table~\ref{tab:appendix_rewards} lists the reward-term weights for each task; blank entries mark terms not used by that task.

\begin{table}[htbp]
    \centering
    \small
    \begin{tabular}{@{}lrrr@{}}
        \toprule
        Reward term & \texttt{in\_palm\_rotate} & \texttt{in\_hand\_repose} & \texttt{screwdriver} \\
        \midrule
        success bonus & $150$ & $500$ & -- \\
        orientation tracking (Gaussian) & -- & $2.0$ & -- \\
        orientation tracking (inverse-$L_2$) & -- & $2.0$ & -- \\
        rotation progress / spin rate & $2.0$ & $0.2$ & $10.0$ \\
        surface-distance reward & $0.1$ & $0.5$ & $1.0$ \\
        off-axis orientation penalty & $-1.0$ & -- & -- \\
        vertical-alignment penalty & -- & -- & $-400$ \\
        object-distance penalty & $-100$ & -- & $-300$ \\
        hand pose-deviation penalty & -- & -- & $-1.0$ \\
        contact-force penalty & $-5.0$ & $-5.0$ & -- \\
        drop / termination penalty & $-100$ & $-100$ & $-100$ \\
        goal-timeout penalty & -- & $-200$ & -- \\
        action-rate penalty & $-5\times10^{-4}$ & $-1\times10^{-5}$ & $-1\times10^{-2}$ \\
        joint-limit penalty & $-40$ & $-20$ & $-40$ \\
        work penalty & $-5\times10^{-3}$ & $-1\times10^{-3}$ & $-1\times10^{-3}$ \\
        \bottomrule
    \end{tabular}
    \vspace{2pt}
    \caption{Reward-term weights per task. Positive weights are shaping rewards; negative weights are penalties. A dash means the term is not used by that task.}
    \label{tab:appendix_rewards}
\end{table}

\paragraph{Domain randomization.}
Every task applies the same actuator randomization, resampled per reset: proportional and derivative gains and motor strength each scaled by $\mathcal{U}[0.95,1.05]$, a position bias in $\pm0.01$ m, a deadband in $[0,0.005]$, gear backlash in $[0,0.01]$ rad, a torque-kick ratio in $[0.9,1.1]$, and per-step torque noise at RFI scale $0.1$.
The task-specific object and disturbance randomization is listed in Table~\ref{tab:appendix_randomization}.

\begin{table}[htbp]
    \centering
    \small
    \begin{tabular}{@{}llll@{}}
        \toprule
        Quantity & \texttt{in\_palm\_rotate} & \texttt{in\_hand\_repose} & \texttt{screwdriver} \\
        \midrule
        object friction ($\times$) & $[0.5,1.5]$ & $[0.3,2.0]$ & $[0.2,1.2]$ \\
        object mass offset (kg) & $[0,0.1]$ & $[0,0.1]$ & $[0,0.2]$ \\
        initial placement $xy$ (m) & $\pm0.02$ & $\pm0.02$ & sampled grasp \\
        initial yaw (rad) & $\pm\pi/12$ & $\pm\pi$ & sampled grasp \\
        external force $x,y$ (N) & $\pm1$ & $\pm1$ & $\pm5$ \\
        external force $z$ (N) & $\pm1$ & $\pm1$ & $[-5,0]$ \\
        force interval (s) & $2$--$4$ & $2$--$4$ & $2$--$4$ \\
        \bottomrule
    \end{tabular}
    \vspace{2pt}
    \caption{Task-specific domain randomization, on top of the shared actuator randomization described in the text. External forces are reapplied at random intervals within the listed range. \texttt{screwdriver} initializes from a set of pre-sampled grasps rather than a randomized free placement.}
    \label{tab:appendix_randomization}
\end{table}

\subsection{Distillation Sweep}

Each task--hand entry in Table~\ref{tab:appendix_distill_matrix} is expanded into the proprioception-only baseline (\texttt{none}) plus the cross product of taxel resolutions (\texttt{low}, \texttt{med}, \texttt{high}), the available placement subsets for that hand, the seven tactile observation types, and both clean and noisy sensor settings.
Type-ablation runs pin resolution to \texttt{med} and use the clean setting; resolution and noise sweeps vary one axis at a time.

\begin{table}[htbp]
    \centering
    \small
    \begin{tabular}{@{}llll@{}}
        \toprule
        Task & Robot & Placement subsets swept & Max student iters \\
        \midrule
        \texttt{in\_palm\_rotate} & \texttt{xhand1} & \texttt{tips}, \texttt{hand} & 6000 \\
        \texttt{screwdriver} & \texttt{xhand1} & \texttt{fingers} & 2000 \\
        \texttt{in\_hand\_repose} & \texttt{xhand1} & \texttt{hand} & 6000 \\
        \texttt{in\_hand\_repose} & \texttt{sharpa} & \texttt{hand} & 6000 \\
        \bottomrule
    \end{tabular}
    \vspace{2pt}
    \caption{Task--hand entries in the distillation sweep.}
    \label{tab:appendix_distill_matrix}
\end{table}

\subsection{Tactile Placement and Probe Counts}

Sensor placement is encoded as a probe asset per (robot, resolution) pair; placement subsets such as \texttt{tips}, \texttt{fingers}, and \texttt{palm} are link filters applied at runtime to the full \texttt{hand} probe set.
Table~\ref{tab:appendix_probe_counts} shows the resulting probe counts.

\begin{table}[htbp]
    \centering
    \small
    \begin{tabular}{@{}llrrrr@{}}
        \toprule
        Robot & Resolution & Hand & Tips & Fingers & Palm \\
        \midrule
        \texttt{xhand1} & \texttt{low}  &  90 &  45 &  78 &  12 \\
        \texttt{xhand1} & \texttt{med}  & 199 & 100 & 164 &  35 \\
        \texttt{xhand1} & \texttt{high} & 667 & 224 & 448 & 219 \\
        \texttt{sharpa} & \texttt{low}  &  98 &  45 &  89 &   9 \\
        \texttt{sharpa} & \texttt{med}  & 206 &  70 & 190 &  16 \\
        \texttt{sharpa} & \texttt{high} & 781 & 245 & 660 & 121 \\
        \bottomrule
    \end{tabular}
    \vspace{2pt}
    \caption{Number of active probes per (robot, resolution) and placement subset used in our experiments. \texttt{tips}, \texttt{fingers}, and \texttt{palm} are link-filtered subsets of the full \texttt{hand} probe set; \texttt{fingers} includes the fingertips.}
    \label{tab:appendix_probe_counts}
\end{table}

\subsection{Sensor Imperfection Parameters in the ``Noisy'' Condition}

Table~\ref{tab:appendix_noise_params} lists the numerical values used in the noisy condition for each observation type, layered on top of the clean parameters from Table~\ref{tab:appendix_sensor_registry}.
The available knobs and their semantics are defined in Section~\ref{app:sensor_noise_model} (Table~\ref{tab:appendix_noise_capabilities}); knobs not listed here are zero or disabled.

\begin{table}[htbp]
    \centering
    \footnotesize
    \setlength{\tabcolsep}{4pt}
    \resizebox{\linewidth}{!}{%
    \begin{tabular}{@{}lcccccccc@{}}
        \toprule
        Type & $\sigma$ & \texttt{rw} & \texttt{quant} & $\rho$ (m) & rel.\ thr.\ (m) & $p_{\mathrm{dead}}$ & $\gamma$ & $(\beta,\tau)$ \\
        \midrule
        \texttt{bool}, \texttt{agg\_bool} & -- & -- & -- & -- & $2\times10^{-4}$ & $0.05$ & -- & -- \\
        \texttt{depth}, \texttt{agg\_force} & $2\times10^{-4}$ m & -- & $10^{-4}$ m & $3\times10^{-4}$ & -- & $0.05$ & $[0.85,1.15]$ & $(0.5,\,0.05)$ \\
        \texttt{force}, \texttt{force\_torque} & $5\times10^{-3}$ N & $10^{-4}$ N & $10^{-2}$ & $3\times10^{-4}$ & -- & $0.05$ & $[0.85,1.15]$ & $(0.5,\,0.05)$ \\
        \texttt{proximity} & $10^{-2}$ & $10^{-3}$ & $10^{-2}$ & $6\times10^{-4}$ & -- & $0.05$ & $[0.85,1.15]$ & $(0.5,\,0.05)$ \\
        \texttt{elastomer} & $10^{-4}$ m & $10^{-5}$ m & $10^{-4}$ m & $3\times10^{-4}$ & -- & $0.05$ & $[0.85,1.15]$ & $(0.5,\,0.05)$ \\
        \bottomrule
    \end{tabular}%
    }
    \vspace{2pt}
    \caption{Sensor imperfection parameters layered on the clean configuration (Table~\ref{tab:appendix_sensor_registry}) in the noisy condition.
    $\sigma$ is additive white noise, \texttt{rw} random-walk drift, \texttt{quant} the quantization step, $\rho$ probe-radius noise, ``rel.\ thr.'' the Schmitt release threshold, $p_{\mathrm{dead}}$ the dead-taxel probability, $\gamma$ the per-reset gain range, and $(\beta,\tau)$ the viscoelastic hysteresis strength and time constant (in seconds).
    Dashes mark knobs that are disabled for that sensor type.}
    \label{tab:appendix_noise_params}
\end{table}

\vspace{2cm}

\section{Sim-to-Real Deployment}
\label{app:sim_to_real}


The real XHand1 SDK reports both a per-taxel raw pressure field and an aggregate contact pressure $f_{\mathrm{calc\_pressure}}=(f_x,f_y,f_z)$ per fingertip.
The hand does therefore expose individual taxel values, but the exact position and response characteristics of each taxel are not documented, so we cannot register the raw field to our simulated probe layout; the most faithful usable signal is the SDK's aggregate contact pressure.
We simulate this as \texttt{agg\_force}, defined by the contact depths along the normal of the surface and scaled by a constant to match XHand1's $f_{\mathrm{calc\_pressure}}$ in axis order and scale.
For \texttt{agg\_bool} we threshold the contact-pressure magnitude per finger,
\begin{equation}
    y^{\mathrm{agg\_bool,real}}=\mathbb{I}
    \left[\lVert f_{\mathrm{calc\_pressure}}\rVert_2>\eta_{\mathrm{real}}\right].
\end{equation}
We deploy the \texttt{in\_palm\_rotate} policy on the real XHand1 and observe 1-2 consecutive rotations before dropping the object.
Note that our teacher policy was deliberately not extensively tuned for robustness and that about 2 consecutive successes is precisely what is observed for fingertips-only \texttt{agg\_bool}.

\clearpage
\section{Temperature Object-Discrimination Experiment}
\label{app:temperature_experiment}

This section details the \texttt{TemperatureGrid} sensor model together with the task and success metric for the temperature experiment of Section~\ref{sec:temperature_experiment} (Fig.~\ref{fig:temperature_experiment}).

\subsection{\texttt{TemperatureGrid} Sensor Model}
\label{app:temperature_grid}

The \texttt{TemperatureGrid} models heat transfer over a voxelized sensor link rather than contact mechanics, returning a temperature field in \celsius.
Each sensor link is voxelized into a grid of $n_x\times n_y\times n_z$ cells from its axis-aligned bounding box, with per-axis voxel size $\Delta = \mathrm{extent}/(n_x,n_y,n_z)$.
The cell temperatures $T$ are advanced each step by diffusion and an internal source, contact conduction, then radiation and convection, with material properties (conductivity $k$, density $\rho$, specific heat $c_p$, emissivity $\varepsilon$, base temperature) assigned per link.
We write $\rho c_p$ for the volumetric heat capacity and $\alpha=k/(\rho c_p)$ for the thermal diffusivity.

\paragraph{Diffusion within the grid.}
We solve the heat equation $\partial T/\partial t=\alpha\nabla^2 T$ with a semi-implicit spectral method.
To impose Neumann (zero-flux) boundaries, the grid is mirror-padded to twice its size along each axis before a real-input FFT, so the discrete spectrum carries the even extension consistent with $\partial T/\partial n=0$ on the walls.
With wavenumber vector $\mathbf{k}$ and $k^2=\lVert\mathbf{k}\rVert^2$, each Fourier mode is updated implicitly,
\begin{equation}
    \hat T \leftarrow \frac{\hat T}{1+\alpha\,\Delta t\,k^2},
\end{equation}
followed by an inverse FFT and a crop back to the original grid; the zero mode is regularized as $k^2\!\leftarrow\!\max(k^2,\epsilon)$.
The update is unconditionally stable, so it does not constrain $\Delta t$ as an explicit diffusion stencil would.

\paragraph{Internal heat source.}
An optional volumetric source field $q$ (in W/m$^2$) is integrated with explicit Euler on the same heat equation,
\begin{equation}
    \Delta T = \Delta t\,\frac{q}{\Delta z\,\rho c_p},
\end{equation}
where dividing by the cell thickness $\Delta z$ converts the surface flux to a volumetric rate before applying the capacity.

\paragraph{Radiation and convection.}
Surface voxels, those on any face of the grid, additionally lose heat to the environment at ambient temperature $T_{\mathrm{amb}}$.
Radiation follows the Stefan--Boltzmann law for a grey body and convection follows Newton's law of cooling,
\begin{equation}
    q_{\mathrm{rad}} = \varepsilon\,\sigma\,(T_K^4 - T_{\mathrm{amb},K}^4),
    \qquad
    q_{\mathrm{conv}} = h\,(T - T_{\mathrm{amb}}),
\end{equation}
where $\sigma=5.670\times10^{-8}$ W/(m$^2\,$K$^4$), $h$ is the convection coefficient, and $T_K=T+273.15$ converts to Kelvin for the radiative term.
The combined loss is applied as $\Delta T = -\Delta t\,(q_{\mathrm{rad}}+q_{\mathrm{conv}})/(\rho c_p)$.

\paragraph{Contact conduction.}
Heat crosses a contact interface by Fourier's law $q=-k\nabla T$.
With the sensor and the contacting body acting as two thermal resistances in series, the interface uses the harmonic-mean conductivity
\begin{equation}
    k_{\mathrm{eff}} = \frac{2\,k_a\,k_b}{k_a+k_b}.
\end{equation}
The conduction length scale is taken as $L=V/A$, one cell's volume over the contact area, which is a heuristic stand-in for the unresolved temperature gradient through the interface.
For a contacting voxel at temperature $T_{\mathrm{cell}}$ opposite a body at $T_{\mathrm{other}}$, the flux, volumetric rate, and temperature change are
\begin{equation}
    \mathrm{flux} = \frac{k_{\mathrm{eff}}\,(T_{\mathrm{other}}-T_{\mathrm{cell}})}{L},
    \qquad
    Q_{\mathrm{vol}} = \frac{\mathrm{flux}\cdot A}{V},
    \qquad
    \Delta T = \Delta t\,\frac{Q_{\mathrm{vol}}}{\rho c_p}.
\end{equation}

\paragraph{Contact area estimation.}
The contact area $A$ is estimated from the contact-manifold points shared by the two links.
We project the points onto their mean contact plane, order them by polar angle about the centroid, and take the polygon area by the shoelace formula.
When fewer than three manifold points are available, each contact falls back to a disk of area $\pi\,\delta$, where $\delta$ is the penetration depth.
In the conduction update the area is additionally floored at $\pi\,d_w\,\delta$, where $d_w$ is a depth-weight scalar (default $1.0$), so that deep point contacts still carry a plausible area.

\paragraph{Sensor element lag.}
The instantaneous cell field $T$ is the physical temperature; the reported measurement $T_{\mathrm{meas}}$ trails it through a first-order RC low-pass filter integrated with forward Euler,
\begin{equation}
    T_{\mathrm{meas}} \leftarrow T_{\mathrm{meas}} + \frac{\Delta t}{\tau}\,(T - T_{\mathrm{meas}}),
\end{equation}
with sensor time constant $\tau$; $\tau\!\le\!0$ disables the lag and reports $T$ directly.
This filter is applied only on the measured readout, modeling the finite thermal response of a physical sensing element, while the ground-truth field is left unfiltered.

\subsection{Task and Success Metric}

\paragraph{Task setup.}
The hand rummages a tilted bin (side $0.5$ m) holding $8$ geometrically identical balls (diameter $8$ cm), one of which is the hot target.
The bin, the hand, and the seven distractor balls start at ambient $T_{\mathrm{amb}}=22\,$\celsius, while the target ball starts at $T_{\mathrm{tgt}}=45\,$\celsius.
The policy observes proprioception and one \texttt{TemperatureGrid} sensor per finger link; it has no other cue to which ball is hot, so the task is only solvable if temperature differences are resolvable through contact.
The hand is reset above the bin and acts at $40$ Hz (a control decimation of $5$ over the $200$ Hz simulation) for episodes of up to $20$ s.

\paragraph{Temperature progress and heat latch.}
For finger sensor $i$ we map its reading to a normalized progress
\begin{equation}
    \pi_i
    =
    \operatorname{clamp}\!\left(
    \frac{T_i - T_{\mathrm{amb}}}{T_{\mathrm{tgt}} - T_{\mathrm{amb}}},\;0,\;1
    \right),
\end{equation}
where $T_i$ is the maximum cell temperature over that sensor's voxels.
A single sticky \emph{heat latch} fires the first time the selected-finger progress reaches $1-m$ with margin $m=0.12$ (equivalently $T_i \ge 42.2\,$\celsius), and remains on for the rest of the episode.
The latch marks the moment the hand has thermally identified the hot ball.

\paragraph{Success metric.}
We score a run by the latched hot-ball contact time: the cumulative seconds for which the heat latch is on \emph{and} at least one fingertip is in contact with the hot ball,
\begin{equation}
    S
    =
    \sum_t \Delta t \;
    \mathbb{I}\!\left[\text{latched}(t)\right]\;
    \mathbb{I}\!\left[\text{fingertip contact with hot ball}(t)\right].
\end{equation}
A configuration counts as a success (a checkmark in Fig.~\ref{fig:temperature_experiment}) when the trained policy sustains this latched contact, $S>0$ across evaluation episodes.
A material configuration whose temperature signal is too weak for the latch to fire, or for which the hand cannot stay in contact once it fires, scores $S\approx 0$ and is marked as a failure.

\clearpage
\section{Contact and Actuation Audio}
\label{app:audio}

Beyond the tactile and thermal sensors above, we develop a proof-of-concept audio pipeline in Genesis that synthesizes both the structure-borne sound of contacts and the airborne noise of joint actuation.
Procedurally generating sound from simulated physics is well established for interactive games and film, where the goal is perceptual plausibility rather than measurement.
We instead propose it as a training signal: a policy that hears its own contacts can infer object properties that the rigid-body solver never resolves.
Audio has already been shown to help contact-rich manipulation~\citep{mejia2024hearing}, and this pipeline makes such a signal available directly inside the simulator.

The motivating observation is one of timescale.
The acoustic signature of a material lives in its vibration modes, which for stiff objects sit in the kilohertz range, whereas the rigid-body solver integrates at a few hundred hertz.
Resolving those vibrations directly would require shrinking the simulation timestep by orders of magnitude, which is infeasible for RL rollouts.
Rather than simulate the vibration, we faithfully approximate it: each physics step emits a short block of audio samples synthesized from the contact state the solver already computes, so the audio carrier sits well above the physics Nyquist frequency while the dynamics step stays cheap.
A physics step of duration $\Delta t$ produces $K$ samples at the sub-step rate $\Delta t_{\mathrm{sub}}=\Delta t/K$, for an audio sample rate $f_s=1/\Delta t_{\mathrm{sub}}$.

\paragraph{Contact audio.}
We use source--filter modal synthesis: the material is a bank of damped resonators, and the contact is the source that excites it.
Each material is a set of modes $\{(f_i, d_i, g_i)\}$ of center frequency, amplitude decay rate, and output gain.
Mode $i$ is a two-pole resonator advanced over the block $k=1,\dots,K$,
\begin{equation}
\begin{aligned}
    y_i[k] &= a_{1,i}\,y_i[k-1] - a_{2,i}\,y_i[k-2] + u_i[k], \\
    a_{1,i} &= 2 r_i\cos(2\pi f_i\,\Delta t_{\mathrm{sub}}),
    \quad
    a_{2,i}=r_i^2,
    \quad
    r_i=e^{-d_i\,\Delta t_{\mathrm{sub}}}.
\end{aligned}
\end{equation}
and the emitted sample is the gain-weighted sum $s[k]=\sum_i g_i\,y_i[k]$.
The excitation $u_i$ is read from the solver contacts on the sensor link: a sharp rise in the normal contact force injects an impulse that pings the modes into an impact ring-down, while sliding injects a velocity- and force-scaled noise source so a scrape colors the same resonances.

\paragraph{Material modes from modal analysis.}
The per-material modes are what make different objects sound different, and they follow from the object's geometry and its isotropic elastic material.
We assemble the linear-elasticity stiffness $K$ and a lumped mass $M$ over a tetrahedral mesh and solve the generalized eigenproblem
\begin{equation}
    K\,\phi_i = \omega_i^2\,M\,\phi_i,
\end{equation}
keeping the non-rigid modes and reading their frequencies $f_i=\omega_i/2\pi$.
Rayleigh damping $C=\alpha M+\beta K$ sets each decay rate $d_i=(\alpha+\beta\,\omega_i^2)/2$, and the gains $g_i$ are the surface normal component of each mode shape $\phi_i$.
Because $K$ scales with Young's modulus and $M$ with density, the modal spectrum is a fingerprint of the material: this is the cue a policy could use to tell a steel box from a wooden one through contact alone, even when the two are dynamically identical to the rigid solver.
The eigensolve is a one-time precompute per object, and hand-tuned material presets are also supported.

\begin{figure}[h]
    \centering
    \includegraphics[width=0.6\linewidth]{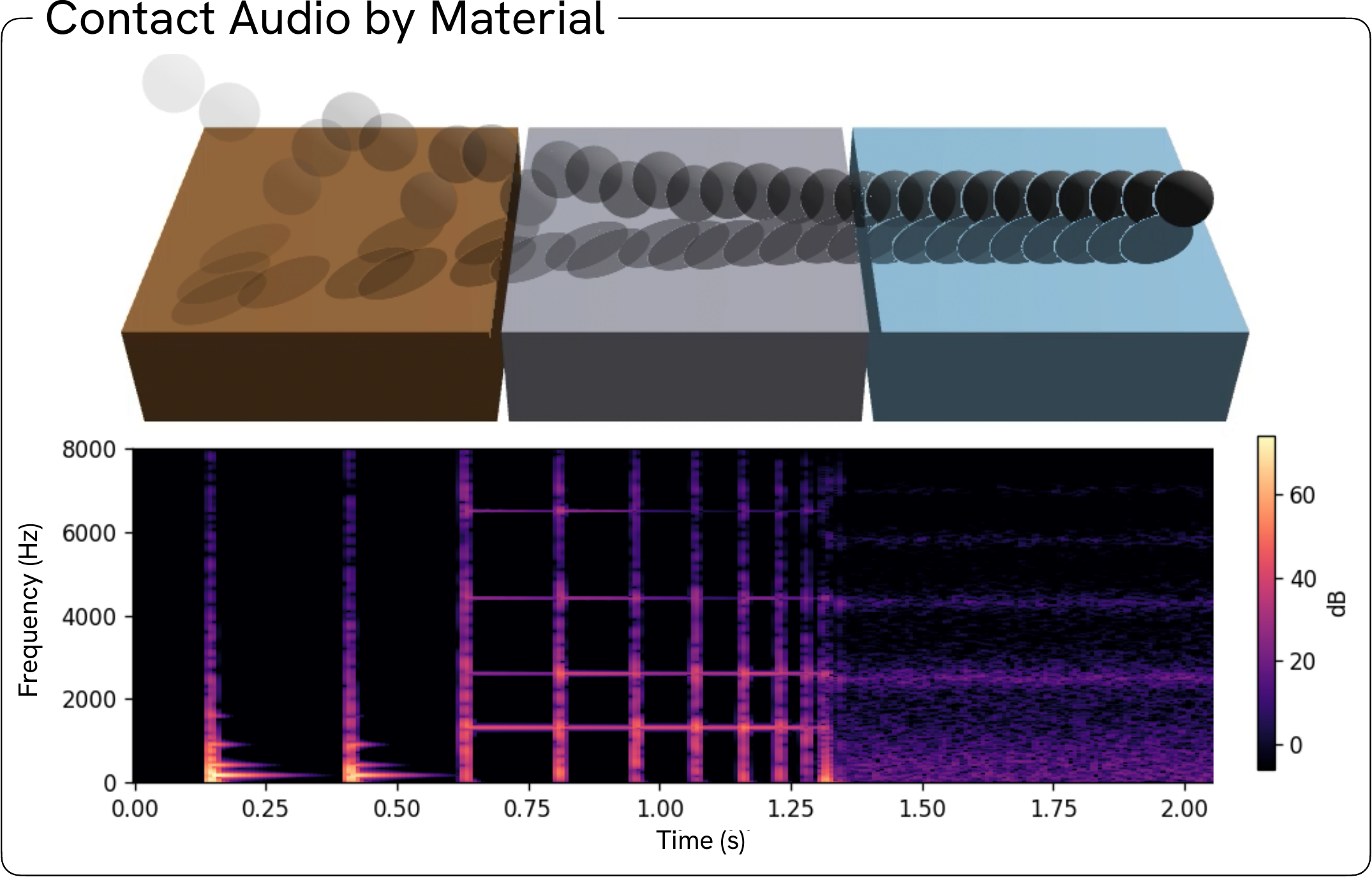} 
    \caption{
    \textbf{Example contact audio by materials.}
     A ball bounces and rolls across a wooden, metallic, and glass box. Although physically identical to the rigid-body solver, the three boxes can be distinguished by their modal spectra.
    }
    \label{fig:audio_materials}
\end{figure}

\paragraph{Actuation audio.}
The actuation source models motor and joint noise with one emitter per actuated DOF, driven by the controller effort $\tau$ and joint speed $\omega$ the solver already exposes.
Its loudness tracks the mechanical load and power,
\begin{equation}
    \ell = \beta_{\mathrm{load}}\,|\tau| + \beta_{\mathrm{pow}}\,|\tau\,\omega|,
\end{equation}
and a velocity-pitched partial bank at fundamental $f_0=\kappa\,|\omega|$ gives the characteristic whine, layered with velocity-scaled friction noise and an idle hum, synthesized with the same two-pole primitives.
A microphone sensor renders what a listener point hears by summing every source over the scene, each attenuated by distance as $1/r^p$ and delayed by $r/c$ for propagation, so contact and actuation sound mix into a single airborne signal.

\paragraph{Toward greater realism.}
This pipeline is a simplified, real-time model, and several improvements can be made to achieve acoustic realism.
Modal synthesis captures the dominant resonances of a struck body, but it omits the frequency- and geometry-dependent radiation efficiency that turns surface vibration into pressure, the room acoustics a real microphone records, and the fine transients of stick--slip friction that our single noise source only coarsely approximates.
The actuation source is likewise a parametric approximation of gear-mesh, bearing, and winding noise rather than a measured motor signature.
Several extensions could narrow this gap.
The dry synthesized signal could be convolved with measured impulse responses of the object and its environment, so it inherits real radiation and reverberation instead of being heard in free field.
Per-material and per-motor recordings could replace or augment the procedural banks, either as sample libraries indexed by contact state and joint speed, or as data to fit the modal frequencies, decays, and gains against rather than tuning them by hand.
A model trained on paired physics-state and audio recordings could learn this mapping directly.
We leave a study of which level of audio fidelity actually benefits policy learning to future work.

\clearpage

\end{document}